# Scaled Hippocampus-inspired Neural Networks on Neuromorphic Memristive Hardware

Joseph A. Kilgore[1], Jeffrey D. Kopsick[2], Zahin Ahmed[1], Giorgio A. Ascoli[2], Gina C. Adam[1]

[1]Department of Electrical and Computer Engineering, George Washington University, Washington, 20052, USA
[2]Center for Neural Informatics, Structures, & Plasticity, George Mason University, Fairfax, 22030, USA

# Abstract

The hippocampus, a key brain region for learning and memory, exhibits rich structural diversity, sparse communication, and robust dynamics with incredible energy efficiency. It offers promising insights for novel computing capabilities, particularly when co-designed with emerging hardware technologies. In this work, we draw inspiration from the rodent CA3 hippocampal subregion to develop the first spiking neural network with neuronal diversity and biologically-realistic resting state dynamics demonstrated on memristor hardware. We propose a network downscaling methodology utilizing a 4-prong objective function and demonstrate a small-scale CA3-inspired network with 179 Izhikevich-modeled neurons, 3 neuronal types and 17,996 synapses with similar resting-state dynamics as the orders-of-magnitude larger full-scale network. The small-scale network is mapped to an FPGA/memristor platform using a greedy algorithm and 18,316 memristors. Benefiting from memristor noise, the hardware implementation shows continuous periodic behavior, outperforming simulated hardware. This work showcases the potential of biologically-realistic algorithms on emerging hardware for neuromorphic computing.

# Main

Biological neural networks with diverse neural populations exhibit robust emergent resting-state dynamics that support critical functionality such as memory consolidation and continual learning, properties that are challenges for artificial intelligence systems[1,2]. Artificial neural networks diverge from this biological baseline with homogenous network architectures while the brain is composed of morphologically and electrophysiologically diverse neuron populations[3]. These diverse subregions are attributed to individual functions, creating a robust overall system. For example, the hippocampus, a central brain region for memory formation and retrieval, and within that is the sub-region Cornu Ammonis 3 (CA3), which has notable capabilities for pattern completion[4]. Implementing realistic networks poses a challenge of parameter choice for biological realism, but large inventories and knowledge bases like Hippocampome.org provide the morphological and electrophysiological data needed to build these models[3,5,6]. Importantly, synaptic parameters are also documented based on a wide collection of the relevant neuroscience literature[7,8], and simulated  incorporating these neuronal and synaptic models to show emergent robust periodic activity[4,9]. Thus, the algorithmic implementation of a data-driven biologically realistic model of a subregion can be analyzed and trained[4,9].

Biological neural networks function on orders of magnitude less energy than artificial neural networks are currently training at[10,11]. For example, GPT-3 was trained with over 1,200MWh, while a brain at 20W would use under 20MWh for an entire lifetime[11]. This has motivated the development of several neuromorphic hardware platforms balancing configurability of traditional computing hardware, with the energy efficiency of the biology[12–17]. Thus, running biologically realistic neural networks on neuromorphic hardware has the potential of extreme energy efficiency while also achieving robustness and accuracy as the biological systems they aim to recreate[18,19]. While platforms like Loihi and Spinnaker have provided large scale networks, on the order of millions of neurons, they lack diverse neuronal models and are less efficient than other emergent analog device technologies[15,20]. For example, full-scale microcortical circuits with 300 million synapses have been simulated with the SpiNNaker system, but utilized homogenous LIF models[21]. Additionally, both of these platforms consume 3 orders of magnitude more energy per synaptic transmission compared to biology[22]. Recent advancements have allowed for customized neuron models with Loihi 2's microcode, these platforms either are entirely digital or mixed-signal with digital transmission of synaptic activity[21,23]. Other platforms, such as BrainScale2, aim towards an analog approach, are yet to incorporate emerging technologies that allow for biomimetic synaptic processing[24].

Memristive devices promise high density in-memory computation with substantial energy savings over traditional computing methods[25,26]. Memristive synaptic connection strength is determined using an analog update based on spike timings of pre- and post-synaptic neurons, with a clear correspondence to the biological base. We use our prototyping platform with 20,000 memristive devices, hosted on a daughterboard with connection to an FPGA[27]. This mixed-signal platform allows for an initial testbed for incorporating noisy analog devices within a spiking neural network. The resulting co-design insights can be used in the future to refine full analog systems and monolithically integrated ASICs to maximize energy efficiency[27,28].

While the CA3 network provides a functionally important and biologically realistic model, its scale is extremely large, 89,226 neurons and 250,078,223 synapses[9,15]. Downscaling this network without losing its capabilities, could provide insights into what neuron and connection types play the most significant role in supporting the overall network activity. Such small-scale biologically-realistic networks could also provide high computational efficiency for edge computing. Their small scale could support direct mappability into novel compact hardware, seeing how hardware non-idealities impact algorithmic perform and allowing for co-design. Thus, it is of high scientific interest to develop a methodology for downscaling these large diverse networks to a smaller scale and demonstrate its implementation on current memristive hardware. While previous work has shown potential methodology for downscaling[29], these methods are not designed for detailed biologically realistic networks to scale sufficiently for prototyping on neuromorphic hardware. Previous downscaling methodologies utilize mathematical constructions designed to maintain neuronal input current levels[29].  Regression fits of biological networks have shown variable exponential functions depending on the parameter (e.g. population size, number of connections, etc.), while previous downscaling work has used linear or squared functions for these parameters[29,30]. Alternatively, other methodologies focus on scaling by abstracting network behavior to population mean-field analysis, which loses the temporal detail and specificity of spiking activity and timing inherently fundamental for STDP and other learning algorithms[31,32].

The novelty of this work resides at the intersection of neuroscience and neuromorphic hardware. We use the model of the rodent CA3 hippocampal subregion as inspiration and downscale it into the first spiking neural network with neuronal diversity and biologically-realistic resting state dynamics demonstrated on memristor hardware. Our novel co-design approach improves upon the biological realism of parameter scaling while also ensuring the network level dynamics remain consistent all the way to the hardware implementation (Fig. 1). Our work is driven by the following open questions, which guide our analysis:

1) What aspects of CA3 network dynamics are impacted as its network size is decreased?
2) Which neuron types and connection types contribute most to stable network dynamics?
3) How do the constraints of memristive devices and prototyping hardware change network dynamics?

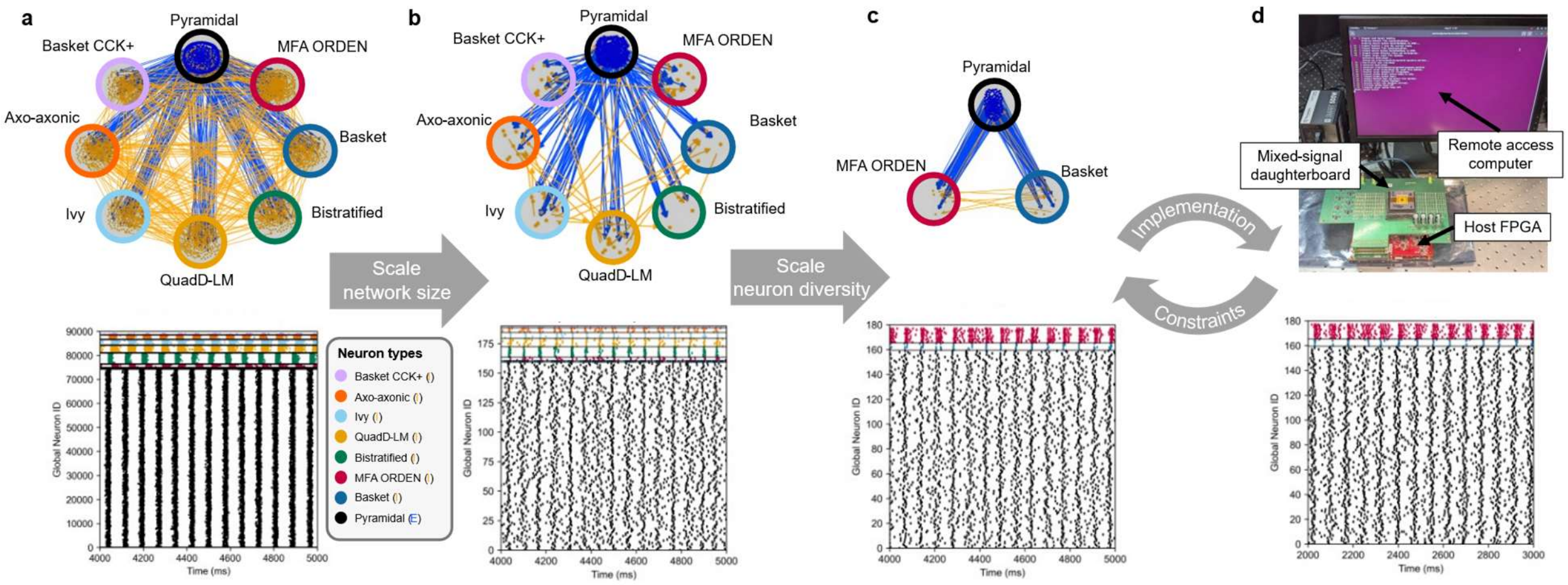

***Figure 1.*** *Overview of network downscaling and tuning towards hardware implementation. a) Full-scale CA3 network simulations show periodic behavior with biologically realistic firing rates*[9]*. b) Utilizing biologically realistic scaling factors and the downscaling methodology based on the 4-prong objective function, a smaller-scale CA3 network variant can be generated to produce similar periodic activity. c) With additional reduction in the diversity of the neuron types, a very small network with <20,000 connections was obtained, still showing resting-state behavior. d) The resulting small network is mapped to an FPGA/20,000 memristor array. Additional tuning based on the hardware constraints allows for generating new small-scale variants that benefit from the memristor noise.*

To answer these questions, we tune a wide variety of network variants at different scales and levels of neuronal diversities based on the full-scale CA3 network using a hyperparameter optimizer and our proposed 4-prong objective function. This novel methodology balances biological realism of network parameters, corresponding architecture and resulting behavior. The full-scale network[9] uses Izhikevich neuronal models of 8 neuron types and biologically-realistic connection probabilities. The resulting emergent activity fits to biological measurements in terms of periodicity and firing rates (Fig. 1a). In this work, we investigate the network scaling for question 1 by reducing the number of neurons using scaling functions fit to biological measurements for the corresponding parameters such as population size and connection probability. Using these base calculations, our hyperparameter optimization focuses on generating continuous periodic behavior at realistic firing rates with minimal changes to projected parameter values (Fig. 1b). To answer question 2, a similar scaling approach is taken, but to investigate networks with reduced neuronal diversity which can show similar resting state dynamics (Fig. 1c). We take the smallest network optimized following hardware constraints and implement it on a physical FPGA/memristive platform to investigate the impact of hardware noise for question 3. We also examine the trends and tradeoffs that occur when downscaling. Finally, we note that this novel hippocampus-inspired network with biologically-realistic resting state dynamics and its implementation on emerging hardware shows potential as a base for next-generation neuromorphic learning systems.

# Impact of Network Size

The proposed method for bio-inspired network downscaling was based on an objective function based on four key prongs: continuous activity, periodic activity, realistic firing rates, and minimal changes from scaled connectivity. Using a spiking network simulator (CARLsim[33]) and a hyperparameter optimization framework (Optuna[34]), we scaled a full-scale rodent CA3 model to reduce its total network population size and its neuronal population diversity, with the goal of maintaining its periodic resting state behavior. Network population size was scaled by a factor of $a^{2/3}$ and total synaptic connections were scaled by a factor of $a$ to follow biological trends observed in the literature[30].The full-scale network was scaled with a value of 0.0001 for $a$ and a reduced diversity to 3 neuron subtypes to produce a network with continuous periodic activity with 17,996 connections. To achieve this, initial downscaling tests utilized values of 0.1, 0.01 and 0.0001 to give a range of networks towards hardware feasibility (Figure 2). Scaled networks shown here are the best network after 2,000 trials, where each trial selects network parameters (e.g. connection probability, background current levels) to optimize network behavior relative to our novel scoring functions.

We observe across scales, from 20,000 to 200 neurons, continuous activity with strong periodic behavior, accounting for 80% of the objective function. Continuous activity is needed to ensure the network activity extends throughout the duration of the trial. Periodic activity is noted for its importance in memory formation and consolidation, both of which are of key importance towards functional usage of a CA3 inspired model[35,36].

The remaining 20% of the score was evenly split between firing rate similarity to the measured biological frequency, and the minimal changes to projected connectivity. As network scale reduced, firing rate scores also reduced from 0.0842 to 0.0000 out of 0.1 possible points, indicating that smaller networks tend toward higher firing rates to maintain continuous periodic activity. Alternatively, the smaller networks had higher change scores, which follows given that the firing rate score could not be optimized, that fewer changes needed to be made to the network connection probabilities. The detailed scoring of 0.1, 0.01, and 0.0001 tuning metrics is shown in Supplementary Figs. 1-5.

Across repeat tunings, we observed that the best individual network from a given tuning session varied in terms of which neurons show biologically realistic firing rates, but no tuning showed all neurons tuning to biologically realistic firing rates at scale (Supplementary Fig. 6).

Beyond the tuning criteria, the Tiesinga-Sejnowski synchrony measure[37,38], based on interspike interval distributions, indicates that subtypes become more synchronous as networks reduce in scale (see Supplementary Table 1). The basket cells showed the largest increase from 0.0373 to 1.0308 from the full-scale to 0.0001 scale network. The trend from most to least synchronous neuron type of the three indicated maintained their order across scales with the pyramidal cells being the least synchronous, and basket cells having the highest synchrony. Results show a decrease in the Gini coefficient with pyramidal cells having the largest drop from over 0.8 in the full-scale and 0.1 scale networks, down to 0.0235 in the 0.0001 network. This reducing indicates that the populations in smaller networks contribute equally to the amount of overall network activity. Given shrinking network populations, this is to be expected and indicates that network activity tends toward even distributions within smaller networks to maintain continuous periodic dynamics.

Utilizing this novel pipeline, we address the first key question of this work: what is the impact of network size on network dynamics? Networks downscaled with a factor of 0.0001 still showed continuous periodic activity, though at elevated firing rates.

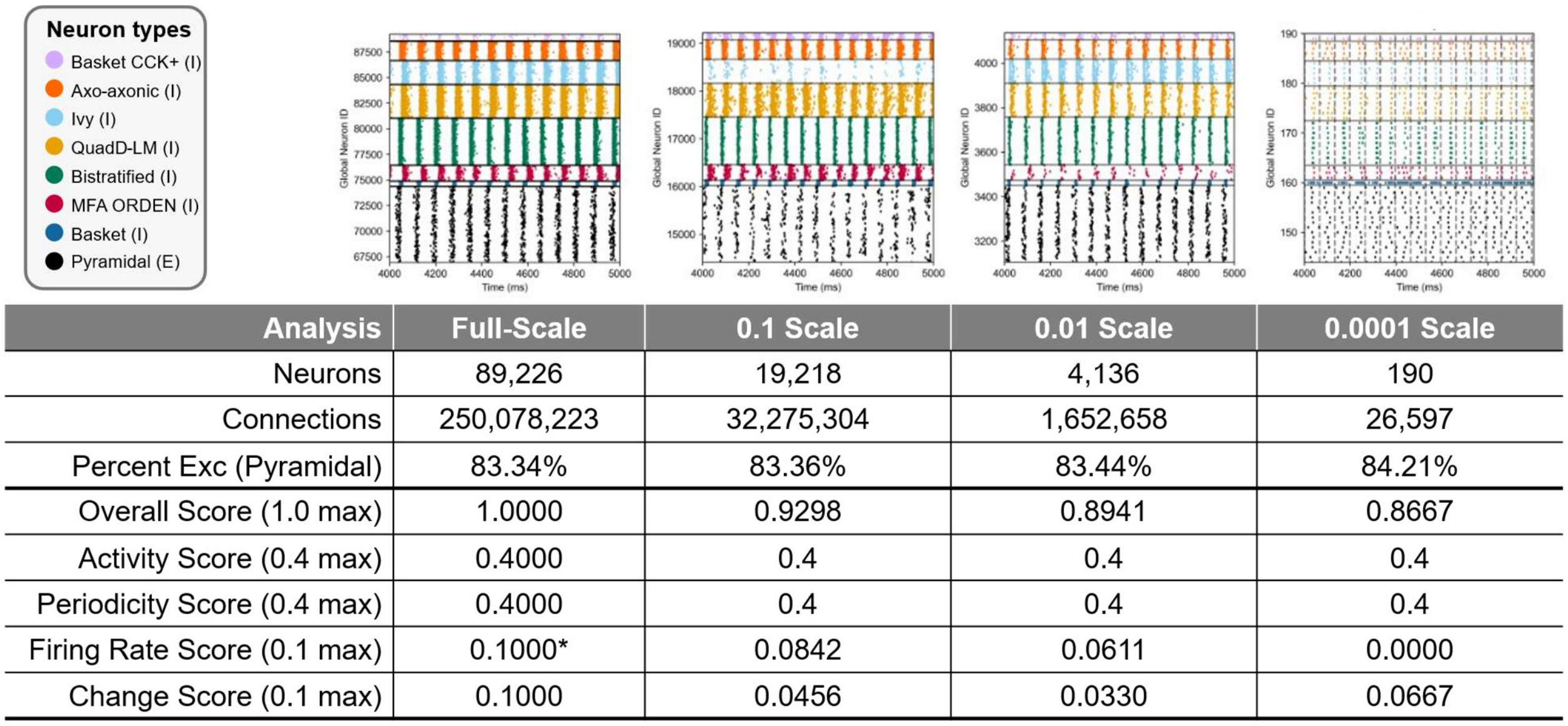

| Analysis | Full-Scale | 0.1 Scale | 0.01 Scale | 0.0001 Scale |
|---|---|---|---|---|
| Neurons | 89,226 | 19,218 | 4,136 | 190 |
| Connections | 250,078,223 | 32,275,304 | 1,652,658 | 26,597 |
| Percent Exc (Pyramidal) | 83.34% | 83.36% | 83.44% | 84.21% |
| Overall Score (1.0 max) | 1.0000 | 0.9298 | 0.8941 | 0.8667 |
| Activity Score (0.4 max) | 0.4000 | 0.4 | 0.4 | 0.4 |
| Periodicity Score (0.4 max) | 0.4000 | 0.4 | 0.4 | 0.4 |
| Firing Rate Score (0.1 max) | 0.1000* | 0.0842 | 0.0611 | 0.0000 |
| Change Score (0.1 max) | 0.1000 | 0.0456 | 0.0330 | 0.0667 |

***Figure 2.*** *Trends across network downscaling. (Top) Network activity from best overall score from 2000 trials (except full-scale model). (Bottom) Network population and connections reduce with biologically realistic scaling factors, creating a stable percent of excitatory neurons in the network.*

*Tuning shows that as scale reduced, the ability to tune also reduced, seen in the reduction in firing rate score and overall objective score. However, all scales could be tuned to exhibit continuous periodic behavior. See Supplemental Table 1 for a detailed expansion including Tiesinga-Sejnowski synchrony scores and Gini coefficients.*
**Note: The firing rate score is based on CARLsim4 full-scale simulation results, making the full-scale firing rate score, by definition, 0.1. This work utilizes the updated CARLsim6, and the resulting firing rates are different, generating a firing rate score of 0.0001.*

# Neuronal and Connection Type Contributions

While network downscaling provides a reduction in neuronal and synaptic populations, the larger network diversity tends toward several groups of a single neuron. Thus, diversity poses additional resource requirements for network implementation. A series of repeat tunings were included to cover a range of diversity from the full 8 neuron types to only 3. The choice for 3 neuron types is based on the analysis of previous work in the literature for their variety in perisomatic and dendritic targeting: pyramidal, basket, and MFA ORDEN[1]. The intermediary steps of 7 to 4 subtypes were determined by removing in order of population, starting with the lowest and finally removing the highest population neuron subtype to get to the 3-subtype network. Each network variation was tuned independently 10 times using the same process as previous downscaling tests.

Scores indicate that across the entire range of subtype variations networks could generate continuous periodic behavior indicated by near perfect 0.4 out of 0.4 scores for activity and periodicity across all repeat trials (Fig. 3a-b). Firing rate scores are on average below 0.01, with 3 and 4 group networks failing to have any firing rates within range to score above 0.0001 out of a possible 0.1 points (Fig. 3c). The change scores, indicating the minimal changes from the project connection probabilities, show that reduced subtype networks require fewer connection probability changes, except for the 3-group network (Fig. 3d).

While the optimized metrics indicate all tested networks created continuous and strongly periodic behavior, untuned metrics will indicate the interplay between individual subtypes. The Tiesinga-Sejnowski synchrony score indicates one neuron type of the 7 inhibitory types, had a substantially higher synchrony compared to remaining neuron types[38] (Fig. 3e & Supplementary Fig. 7). Spiking activity, counted as average spikes across the last second of the simulation, indicates reduced inhibitory activity with reduced subtypes, as expected (Fig. 3f). However, reduced inhibitory activity led to an increased excitatory activity, trending toward >90% excitatory activity in the 3-subtype network variant vs. <80% in a network with full neuronal diversity. Excitatory and inhibitory balance is important for network training and behavior, so it should be considered when reducing the network complexity[39]. While excitatory and inhibitory activity diverge as network complexity reduces, the overall trend of total network activity reduces with both complexity and size. The non-monotonic nature of the activity across different diversity levels may indicate that particular neuron types, or combinations, may allow for more activity than a pure linear regression would predict (Fig. 3g). As an example, the reduction from 8 neuron types to 7 removes the Basket CCK+ subtype and causes a corresponding overall activity drop. The subsequent removed subtype, Axo-axonic, causes the network to go back up in overall activity, noted previously in the increase in overall excitatory activity increasing. This indicates the potential for different neuron types, in this example the Basket CCK+, to have a stronger role in reducing overall excitatory activity. Finally, synaptic transmission is also not linear, further indicating that particular neuron types, once removed, cause greater changes in network dynamics compared to other neuron types (Fig. 3h). However, regardless of the combination tested, networks could still create continuous periodic activity, often with minimal changes to the projected connection probabilities.

To provide a balance of diversity and scale, an additional hyperparameter was added to allow for population scaling of inhibitory neuron types. Utilizing the smallest variant, 3-subtypes at 0.0001 scale, an additional tuning generated the final network with 160 pyramidal cells, 12 MFA ORDEN cells, and 7 basket cells with a total of nearly 18,000 synapses (Fig. 5a-b). The resulting allowance for increased inhibitory populations, allowed for what was previously a single basket cell to become a population, and created an 89:11 excitatory:inhibitory ratio of neurons. Additionally, the increased populations allowed for lower firing rates, generating a 3-group network with a firing rate score of 0.33 (Fig. 5e & Supplementary Table 1).

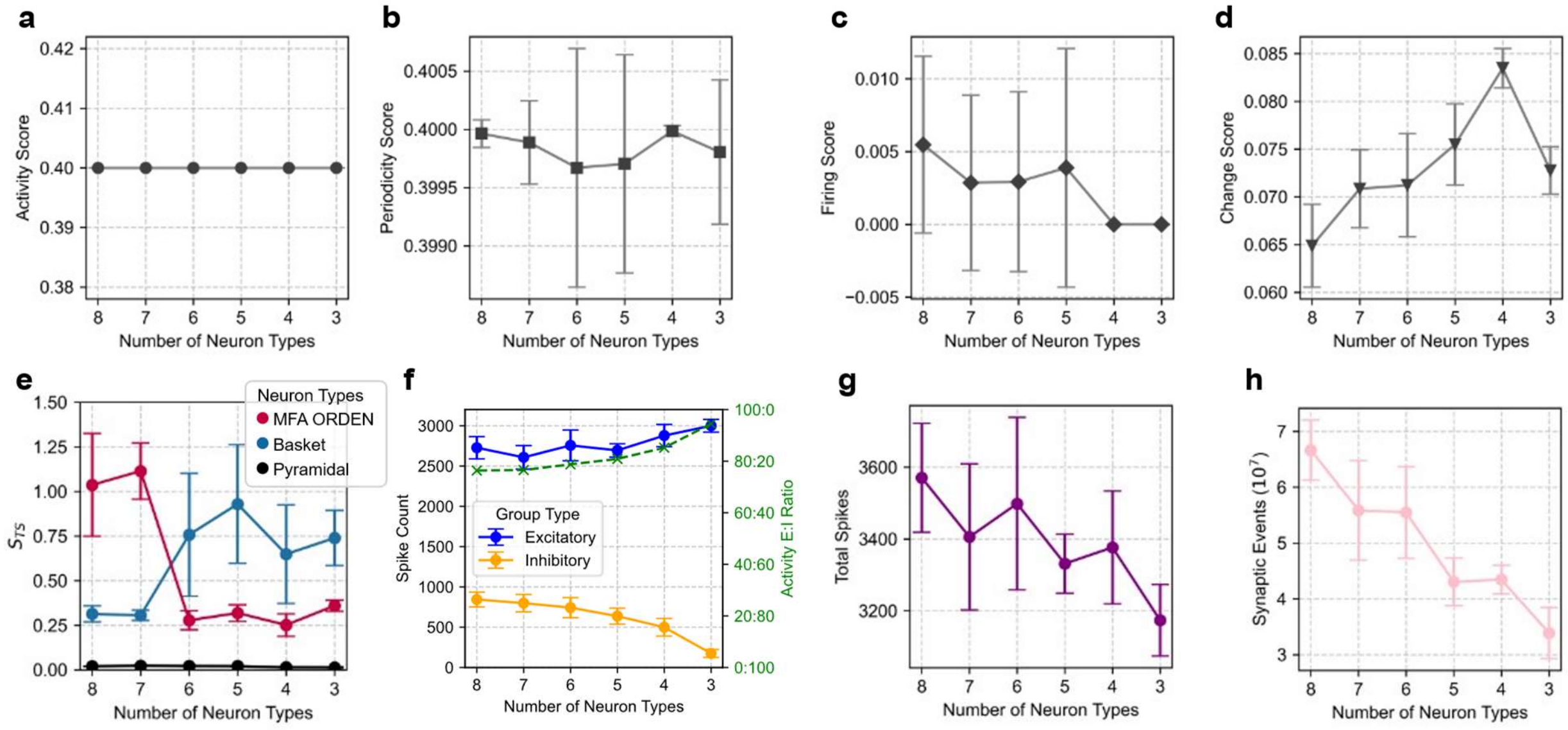

***Figure 3.*** *Comparison of network diversity from a full-diversity network (8 neuron types) to an archetype network (3 neuron types) with 10 separately tuned networks at each network configuration. a) Activity score, indicating the level of continuous activity with a max score of 0.4 which all 60 tuned networks scored. b) Periodicity score, measuring the strength of oscillations in the network. Nearly all networks scored a max of 0.4. c) Firing rate score shows the ability of tuned networks to mirror biologically realistic firing rates with a max of 0.1. Networks with 3 or 4 neuron types score < 0.00001 indicating that reduction to this scale produces a network that is unable to maintain continuous periodic activity with biologically realistic firing rates. d) Change score indicates the similarity to projected connectivity the tuned network exhibited. e) The Tiesinga-Sejnowski synchrony indicates that neuron types would vary across the diversity range, with Pyramidal cells being the only consistently low synchrony neuron type. f) Activity in the network, summed over the last 1 second of simulation time, trends toward proportionally higher excitatory activity with reduced network complexity. g) Excitatory and inhibitory neurons of the last*

*1 second were summed to provide total spikes across the network which follows a reduction with reduced network diversity. h) Mean active synapses was measured by comparing calculating the synaptic connections stimulated by all spikes in the last 1 second of simulation providing a measure of synaptic events that occurred in the network.*

This analysis addresses the second key question of this work: what neural subtypes and connections create stable small-scale network dynamics? We demonstrated that networks from 8 neuron types and 51 connection types down to networks with 3 neuron types and 9 connection types maintained continuous periodic activity. However, overall network activity was impacted in a non-linear fashion by the reduction in diversity. This indicates that each neuron type had distinct impact on network activity, e.g. as noted by the fact that removing Basket CCK+ neurons vs. Axo-axonic neurons changes the network activity in opposite ways. Together, these provide the basis for generating novel small-scale CA3 inspired networks at a sufficiently small scale for our memristive prototyping platform.

# Implementation on Memristive Hardware

Incorporation of memristive devices as synaptic transmission for neurons was successfully implemented within our prototyping platform. A complete overview is shown in Figure 4, with additional implementation details provided in the corresponding methods subsections.

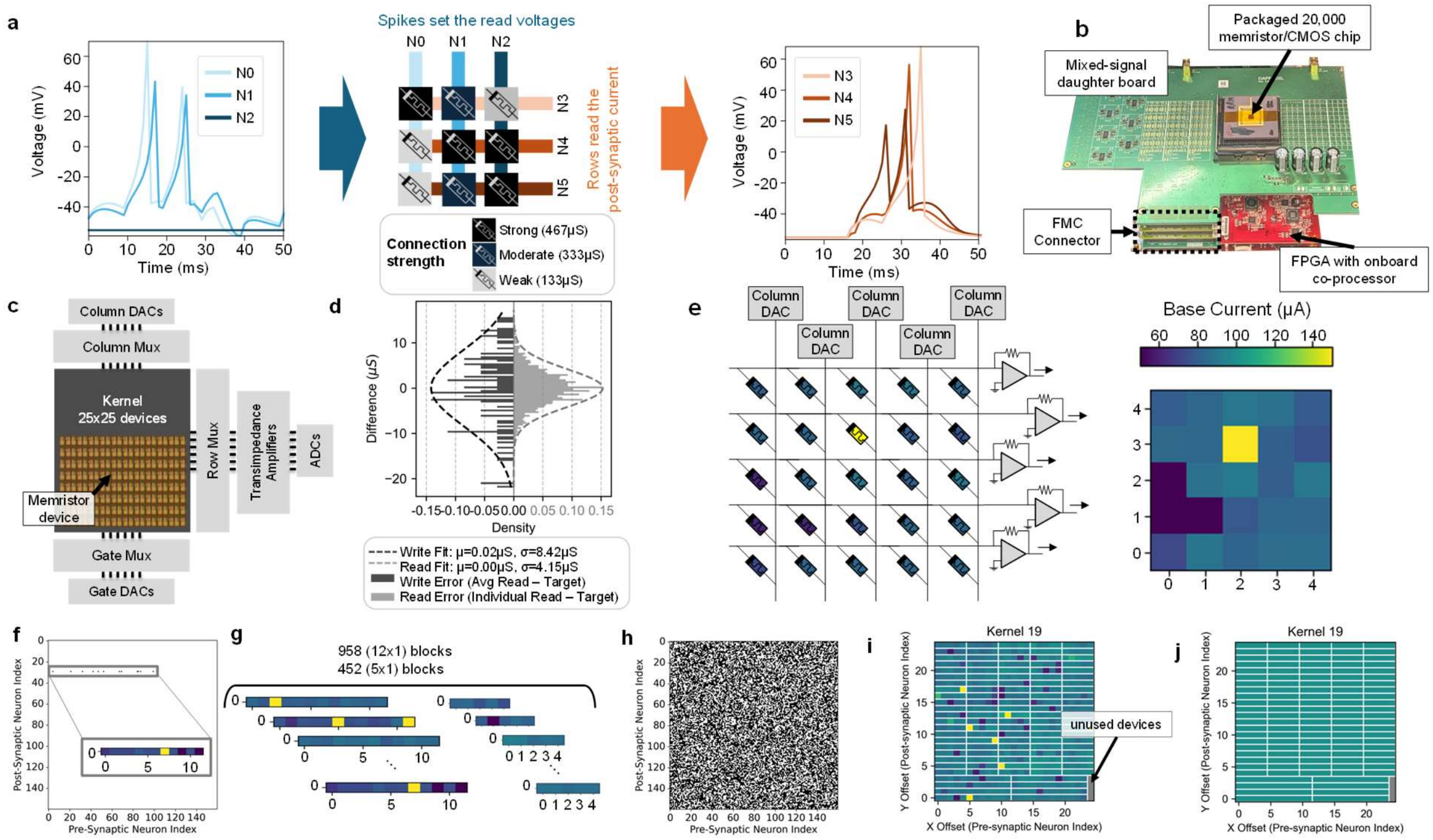

***Figure 4.*** *Hardware implementation on our memristor prototyping platform. a) Example group-to-group synaptic connection processed through a memristor array. Each timestep queries the memristive array, utilizing read voltages to indicate pre-synaptic spikes. The outgoing current is*

*converted to post-synaptic current for the receiving neurons. b) The hardware prototyping platform consists of an FPGA, mixed-signal daughterboard, and the memristor crossbar array. c) Architecture of the prototyping platform, showing how the connection matrix of the network is mapped to kernels (25x25 device blocks), which are connected to DACs and ADCs for communication with the FPGA d) Memristor write error measured across 90 target conductance levels and device read error measured across 5,760 read operations. e) Example of a 5x5 block of memristor devices for synaptic mapping. Memristor devices stuck in high conductance (shown in yellow) pose additional considerations for hardware mapping with the exact current measurements using a 0.3V read voltage. f) First block selection in the pyramidal-to-pyramidal connection. This is a 12x1 set of connections with 12 random presynaptic neurons and 1 random postsynaptic neuron. g) These connections are randomly sampled in groups of 12x1 until a block with minimal overlap is not found within 5000 random resamples. The remaining blocks are sampled in groups of 5x1, producing a total of 1410 total blocks. h) The final pyramidal-to-pyramidal connection probability is 53.73% with no connections overlapped using these block sizes. i) Example of network mapping on kernel 19 shows the packing of 12x1 and 5x1 blocks. The plots of the 31 kernels used in the hardware mapping are shown in Supplementary Fig. 14. j) The same kernel 19, but with homogeneous simulated devices.*

The reduced diversity allowed for connection matrices that reached the 20,000-synapse threshold for implementation in our memristive hardware. Utilizing the 3-scale network with increased MFA ORDEN and Basket cell types, the resulting network architecture, shown in Fig. 5a-b, was passed through a separate translation algorithm to generate blocks of contiguous synaptic connections, allowing for fewer device read operations. Trials were conducted with simulated devices (non-noisy) as well as experimental results on physical memristive devices within the mixed-signal prototyping platform.

The simulated network again produced continuous periodic behavior at each step of the process, with activity scores and periodicity scores dropping from CARLsim to simulated devices (Fig. 5c, 5e). This is expected due to the reduced variability in synaptic fan-out caused by the mapping process. Blocking was required to reduce the number of read operations which are time intensive. The final physical hardware implementation produced superior behavior based on tuning metrics compared to the simulated devices (Fig. 5a-e). The noise introduced from analog devices, both in terms of read and write error, countered this reduced structural diversity to improve across all tuned metrics. Additionally, the overall trends for interspike intervals across subtypes and simulators remain relatively constant, with the notable exception of the substantial proportion of basket cells tending toward a consistent 75 ms ISI in CARLsim, which is lost when translating to memristive devices either simulated or physical (Fig. 5f).

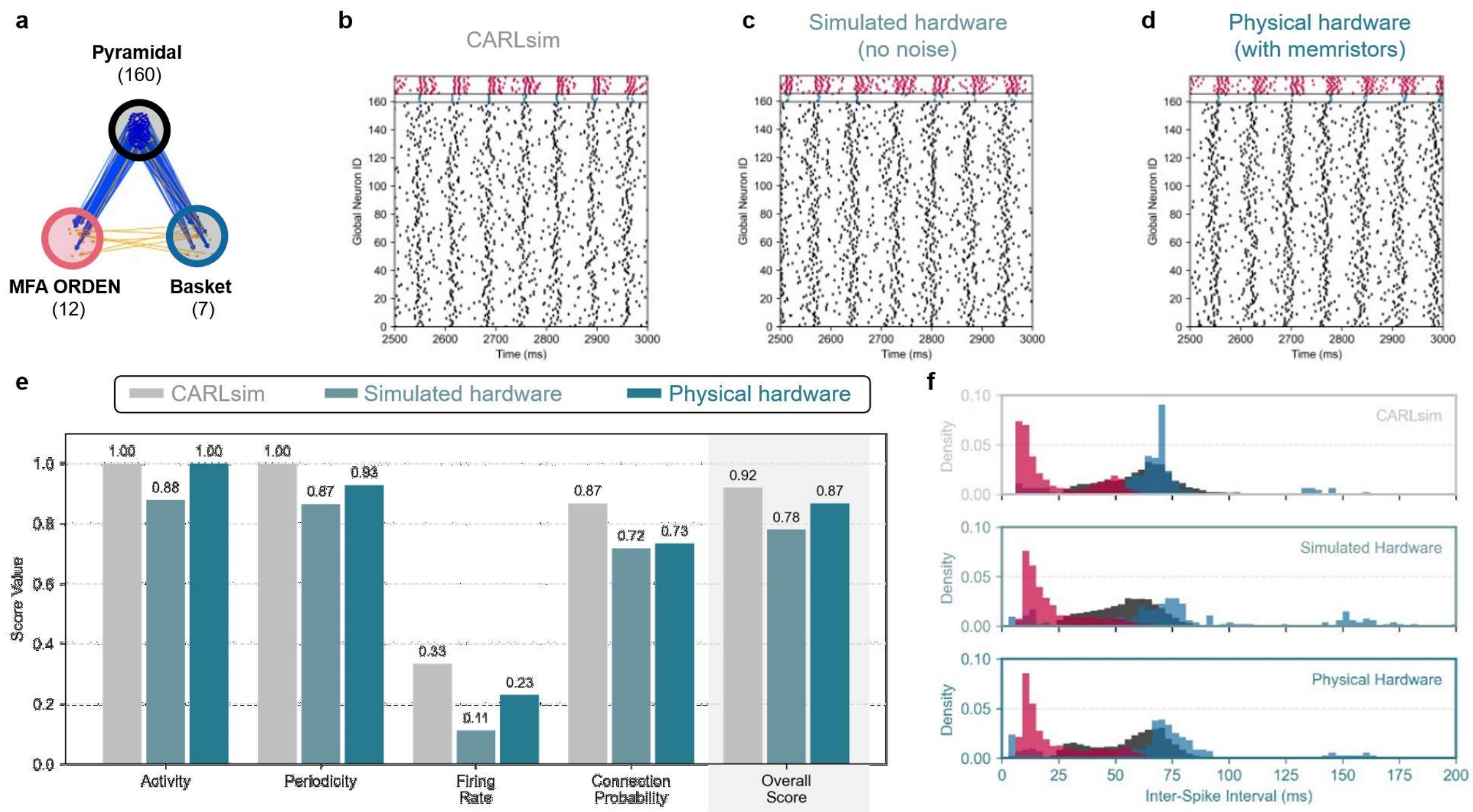

***Figure 5.*** *Comparison of network activity in simulation and physical hardware. a) Network architecture after tuning allowing for controlled population multipliers to inhibitory groups to allow for adjustments to account for reduced inhibitory diversity. b) Raster plot of network activity simulated with tuned network parameters and randomly generated connectivity in CARLsim6. c) This network was translated to hardware, requiring a mapping of connections to an efficient set of device blocks and simulated again using virtual non-noisy devices. d) The network translated with physical devices for synaptic connections. e) Comparison of simulators and hardware network tuning objective scores. f) Comparison of interspike intervals by simulator/hardware and subtype.*

Beyond the tuning metrics, additional non-tuned metrics describe the activity inequality and synchrony. The Gini coefficient and Tiesinga-Sejnowski synchrony show nearly identical trends (Fig. 6a-b and Supplementary Figs. 8-9). Little change is observed in the pyramidal cell population in either metric. The basket cells show a substantial increase in inequality (through the Gini coefficient) which indicates the synaptic blocking process has a significant impact on the relative activity that is transmitted to the small basket cell population. However, the noise of physical memristive devices, as previously noted across tuning metrics, brings the coefficient back down to near CARLsim levels. Again, this same trend in basket cells is noted with synchrony, with an increase in basket cell synchrony on simulated devices and near baseline CARLsim synchrony with physical devices. The final 1000 ms of activity was recorded and shown to compare to the previous results in Figure 3f (Fig. 6c), noting that the allowance for hyperparameter optimization of inhibitory populations leads to a network with an 80-20 split of excitatory and inhibitory activity.

To estimate preliminary power consumption metrics of the CA3-inspired network on this first generation memristive device platform, the sample physical device parameters are shown in Figure 6d. The values show the target conductance level of devices used in the full-scale simulation, and the potential lower conductance values. The chosen target conductance (333 µS) is based on a value near the middle of the conductance range, since future work may incorporate on-device learning (Fig. 6d). The generated currents are then converted with an optimized multiplicative factor when generating current for the Izhikevich models. Because a device read is synonymous with a synaptic transmission, this metric provides a parallel to power consumption. Across the 3,000ms trial, the pyramidal cells generated 71.7% of all network transmissions. Across the three neuron types, the MFA ORDEN cells had the greatest synaptic events per neuron at an average of 7,887 per neuron (Fig. 6e-f).

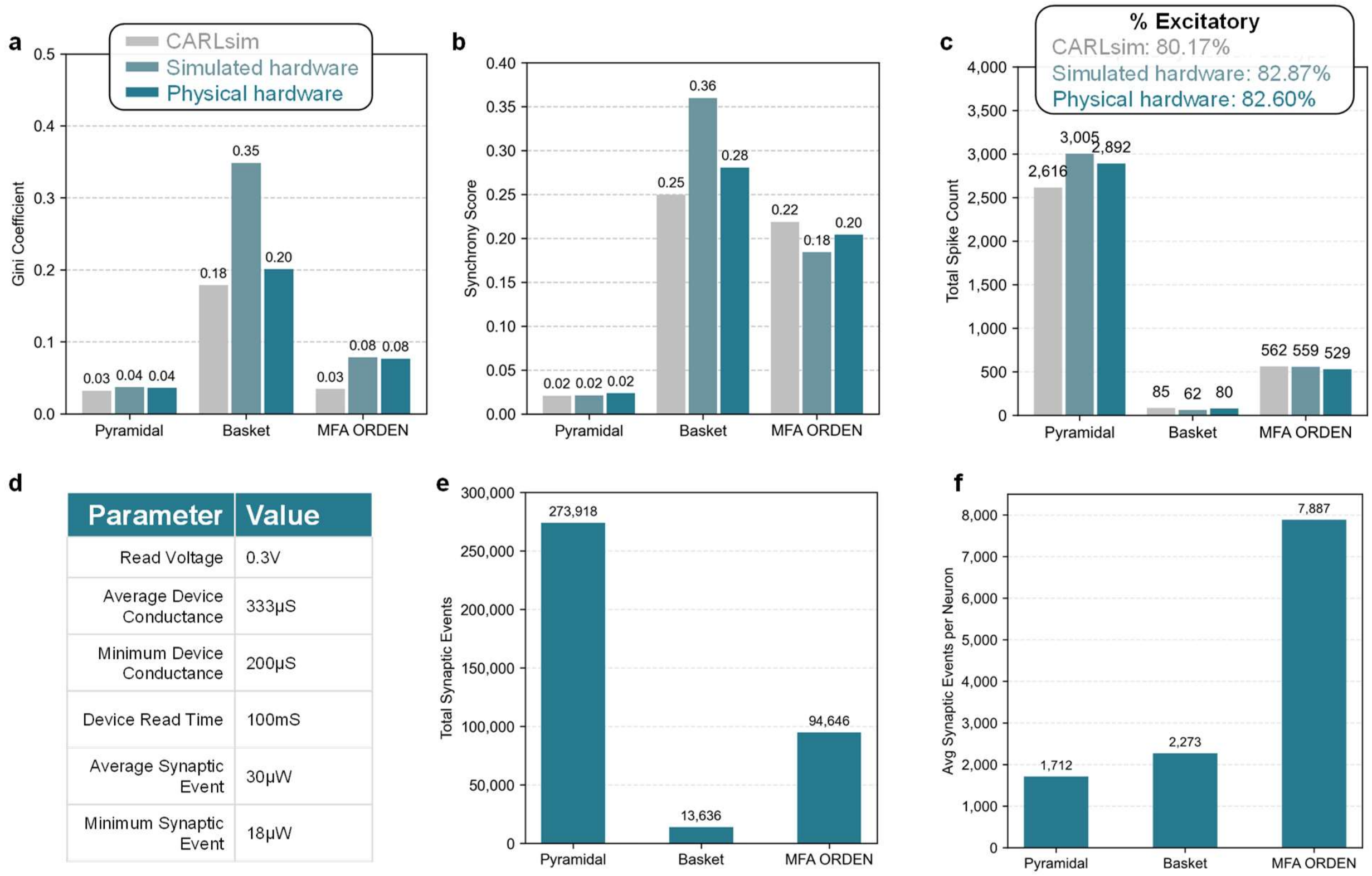

| Parameter | Value |
|---|---|
| Read Voltage | 0.3V |
| Average Device Conductance | 333µS |
| Minimum Device Conductance | 200µS |
| Device Read Time | 100mS |
| Average Synaptic Event | 30µW |
| Minimum Synaptic Event | 18µW |

***Figure 6.*** *Beyond tuning metrics of memristive device networks. a) Comparison of the inequality of neuron spiking activity, Gini coefficient, by subtype and simulator. b) Tiesinga-Sejnowski synchrony scores across subtypes and simulators. c) Totally spiking activity by subtype, with the memristive device networks increasing the overall excitatory inhibitory ratio of activity from 80% to 83%. d) Physical memristor parameters and average power consumption per device read. The average device conductance is based on the target level for all devices used in this simulation, however since conductance isn't designed to change during simulation, a lower device conductance could have been used to lower overall power consumption. e) Utilizing the power consumption metrics derived from device conductance read voltage, the total number of device read operations, noted as total active synapses, is shown both as the raw number of events and the estimated power consumption (excluding DACs, ADCs, and other IO). f) The synaptic events are divided by population size to determine power usage per neuron by subtype.*

Turning to the final question we aim to address in this work, we show that our prototyping hardware provides sufficient capabilities to produce continuous periodic behavior through noisy memristive synaptic connections. Given the spatial constraints within kernels with 25x25 devices, the exact connection matrix had to be modified through a greedy algorithm to provide efficient device usage. Efficient mapping led to reduced overall diversity in network connectivity, due to connections being selected in equal size groups instead of randomly. This reduction in synaptic diversity directly impacts the network ability to produce biologically realistic behavior, apparent in the hardware simulation with reduced scoring across all tuned metrics.

However, the final translation to physical hardware benefited from the inherent noise in the memristors, DACs, and ADCs to counter the reduced structural noise introduced through connection blocking. All scoring metrics improved in the hardware experiment which reached similar network resting-state behavior with the algorithmic demonstration. The hardware implementation ran for a total of 10 days due to 100ms delays between read operation, lack of parallelization across kernels, and data saving. We note that this overall solution does have considerable overhead due to the DACs and ADCs and no kernel level parallelization. Kernel level parallelization is limited by DACs and ADCs, but further software developments should support parallel operations. These results give guidance for future developments in fully analog solutions as well as prototyping hardware that can support larger implementations of multi-subregional models.

# Conclusions

This work introduced a complete pipeline from a large-scale neuroscientific model to small-scale hardware implementation and demonstrated resting state behavior in memristor-mapped small-scale CA3-inspired spiking neural networks. The proposed methodology generated novel small-scale networks with continuous periodic activity. Periodic activity, specifically in the form of synchronous oscillations, plays a well-established role in memory consolidation and sequence replay[35,36], and is thus critical for small-scale networks to exhibit for future functional behavior such as continual learning and memory storage. Our analysis provides considerations when adapting this methodology to other types of networks, e.g. higher firing rates at smaller scales. This approach also shows how a biologically inspired neural network with neuronal diversity can be translated to a hardware prototype and what constraints might have to be considered, particularly for emergent technologies.

The novel pipeline for network tuning is designed to be model agnostic, based explicitly on spiking activity, and allows for tuning of other sub-regions, multi-regions, or swapping the underlying model, simulator, and hardware substrate. These results show the first implementation of CA3 resting-state dynamics mapped to 18,316 memristive devices and lay the groundwork for future neuromorphic systems inspired by the hippocampus with capabilities for continual learning and large memory storage.

# Methods

## Tuning Pipeline Overview

The methodology is based on four proposed metrics and is tested on a biologically realistic CA3 network as a representative example. We utilize Optuna, a hyperparameter optimization framework[2], to control connection probability and input current levels, and optimize four criteria. However, it is worth highlighting that the proposed methodology could be applicable broadly for downscaling various biologically realistic networks inspired by different neural systems in an automated fashion due to the analysis throughout tuning being entirely spike based. The key issue that this work addresses is the need for generating biologically realistic dynamics as network scale reduces, while maintaining as much biological realism as possible. Previous manual tuning is prohibitively slow given the number of potential parameters in a biologically realistic network, well over 100 in the mouse CA3 model example. We propose a method that combines overarching network behavior, continuous and periodic spiking, with biologically realistic targets, firing rates and synaptic connection probabilities, with an automated system for parameter tuning and network analysis. With this, networks of any scale and region or subregion can be generated given a base set of network parameters. Additionally, the tuning itself is separate from the simulator, and thus any underlying model/simulator combination could be utilized. The overall pipeline is broken into 3 key pieces. (1) the definition of a baseline model, in this work we utilize a full-scale rodent CA3[9]. (2) The scaling process from the initially defined network to a user-defined scale. (3) The iterative process of parameter choice and simulation scoring. These three pieces define an entire pipeline for network tuning and are discussed in detail for our specific use case in the following sub-sections. This tuning pipeline is then connected with a mapping algorithm to the hardware ecosystem to create the complete downscaling to hardware implementation (Supplementary Fig. 10).

### CA3 Network and Simulation Protocol

The parameters for the full-scale network utilized as a benchmark in this work are from Kopsick et al. creating a robust real-scale model of the mouse CA3 with 8 neuron types[9]. These 8 include excitatory Pyramidal cells and 7 inhibitory interneuron types: Axo-axonic, Basket, Basket CCK+, Bistratified, Ivy, Mossy Fiber-Associated ORDEN (MFA ORDEN), and QuadD-LM cells. They were chosen to represent a range of supertypes categorized in Hippocampome.org[3,9]. The model also includes 51 distinct connections between directed pairs of pre-and post-synaptic neuron types, each with specific connection probability and synaptic conductance. Every neuron type has its characteristic firing fitted by a biologically realistic 9-parameter Izhikevich model that can be easily implemented into the spiking neural network simulator CARLsim6, see Supplementary Tables 2-4 for network parameters[6,33,40].

To stimulate activity in the network, an input current protocol is utilized. The network is stimulated at a high current for the first 10 ms to generate activity in the 1/10$^{th}$ of pyramidal cells, along with a background current provided to all cell types for the duration of the simulation. All runs continued for a total of five seconds unless firing rates were measured over 250Hz upon which the run would end immediately and automatically received an overall objective score of 0. This design drastically increases efficiency since computational complexity with CARLsim increases with number of spikes.

The choice of network and simulator here are based on the availability of substantial biological realism of the choice network and it's existing implementation on CARLsim[9]. However, if the parameters are explicitly defined and provided to the hyperparameter optimization framework, the connection between tuning algorithm and network simulator is abstracted to a single output data file to allow for future expansion to other simulators and architectures. For additional details on pipeline architecture, please see Supplementary Fig. 10.

## Network Scaling

Network parameters are scaled based off a log domain scaling specific for each parameter following (1):

$$y = a^x y_0 \quad (1)$$

Where $y_0$ is the parameter being scaled by a factor $a$, and $x$ is the parameter specific scaling factor. Accordingly, the numbers of neurons and synapses in the network can be scaled, providing the scaled biological values. These two factors are then used to calculate the connection probability. The literature[30] states that the value of $x$ for neuron population scaling is 2/3 and the value for scaling the number of synapses is 1. Based on these two values, the connection probability between neuron groups can be calculated. These calculated connection probabilities are used as the base scaled connection probability value for network scoring.

## Scoring Overview

We propose a methodology for deriving and analyzing CA3 network variants at various sizes and comparing their characteristic activity with the full-scale network. We define four target criteria for network variants as follows:

1. Scaled networks should display continual activity beyond the transient stimulation and throughout the duration of the simulation.
2. Scaled networks should show oscillatory dynamics.
3. Mean firing rates of each neuron type in the scaled network should match those of the full-scale network.
4. If changes in connection probability are necessary for the previous three points, then the changes should be as minimal as possible.

Each of these criteria are evaluated on a scale of zero to one and weighed to derive the objective equation for Optuna (2).

$$O(N) = 0.4A(N) + 0.4P(N) + 0.1F(N) + 0.1C(N) \quad (2)$$

Where $O(N)$ is the objective value of the network, $A(N)$ is the activity score, $P(N)$ is the periodicity score, $F(N)$ is the firing rate score, and $C(N)$ is the change in connection probability score. Each of these scores individually range from zero to one, so the continual network activity and periodicity of the network make up 40% of the overall score each, and the firing rates and connection probability scores together make up 20% of the overall objective function. The overall objective function, while being ranged 0 to 1, is made up of these 4 sub-functions, each also ranging from scores 0 to 1. Users may decide which attribute is more important in their task and adjust weights accordingly.

## Activity Score

Criteria 1 emphasizes continuous network activity, and the activity score aims to score a value of 0 for inactivity and a value of 1 for continuous activity. To do this, the final 1 second of simulated time is extracted and binned into 66ms increments. This bin size was chosen based on the activity frequency of the full-scale CA3 network, which shows strong periodic activity at ~16Hz. Hence, the bin size reflects a lower frequency of 15Hz designed to ensure each bin will capture the activity mirrored in the full-scale network[9]. Activity for each subtype is binned separately, and finally we check how many bins have at least 1 spike in them.

$$A(N) = \left( \frac{1}{(\#Groups) * (\#Bins)} \sum_{g=0}^{\#\,Groups} \sum_{b=0}^{\#\,Bins} S_{g,b} \right)^2 \tag{2}$$

Where $S_{g,b}$ is zero if there are no spikes in that subtypes bin, or one if there is at least one spike in neuron group $g$ during time bin $b$. To improve the hyperparameter optimization, additional emphasis was added for activity to occur in all bins scored. To this aim, we square the result yielding the final activity function (2).

In this function, all bins and neuron types are equally weighted. Given the full-scale network has 8 neuron types, and there are 15 bins for each type, that gives a total of 120 bins to score. An example scoring is shown in Supplementary Fig. 1.

The choice 15Hz binning frequency should be sufficient in most applications but should be considered depending on the target frequency of network oscillations. Notably, a tradeoff between size and firing rates, with smaller networks requiring higher firing frequencies to maintain activity, and larger networks having lower firing rates with more variability, balance to allow for 15Hz binning to be applicable across the range of tested network scales.

## Periodicity Score

To score criteria 2, a similar process to the full-scale CA3 network frequency analysis in Kopsick et al. was used but expanded by scoring the periodicity of the network quantitatively on a scale from zero to one[9]. By measuring a first approximation of the local field potential (LFP), and utilizing a Fourier transform, analysis in the frequency domain showed frequencies with substantially higher power than baseline. Notably ~16Hz beta oscillations are prevalent as previously shown, as well as other frequencies across different oscillation ranges[9].

The approximate LFP is defined as the average voltage of a particular neuron group. For computational and memory efficiency this is the average voltage of a subset of the Pyramidal cells. For frequency analysis, the first 3 seconds of activity are skipped to avoid the initial transient period, though this varies depending on the availability of data from the simulation platform, with physical hardware producing scores based on less data due to physical constraints. Then a power spectral density of the signal was created by taking the absolute value of the squared Fast Fourier Transform. This power spectral density was smoothed with a moving average and fit with a linear regression in the log-log domain. This fit created the baseline across the frequency spectrum, from which a ratio of network frequency power and the baseline frequency power was calculated. Wherein the resultant output, peaks are based relative to their baseline power across the entire frequency spectrum. This peak to baseline ratio emphasizes the strength of the peak of activity, while providing a base line of random activity having no strong peak and thus easily measuring a weak versus strong periodic behavior. This ratio is used for the periodicity score

$$P_{range}(N) = f(x) = \begin{cases} 0.0, & P_{N,f} < P_{B,f} \\ 0.1 * \left(\frac{P_{N,f}}{P_{B,f}} - 1\right), & P_{N,f} < 11 * P_{B,f} \\ 1.0, & P_{N,f} \geq 11 * P_{B,f} \end{cases} \quad (3)$$

Where $P_{N,f}$ is the power of the network LFP at frequency $f$, and $P_{B,f}$ is the power of the log-log fit at the same frequency. The frequency range was broken into three oscillation ranges, based on known biological relevance. These are the theta (4-12Hz), beta (13-25Hz), and gamma (25-100Hz) rhythm ranges[41,42]. Everything below 11x the base power is given a linear scaling starting at 1x base power. All peak power above 11x the base power is given a max score of 1.0. The peak score from each range is summed below to give the final periodicity score (Supplementary Fig. 2).

The beta range receives the largest weight due to its prominence in the large-scale network resting state[9]. The theta and gamma ranges receive lower weighting, but it is possible that smaller networks will likely have stronger exhibition of gamma rhythms. Additionally, other subregions may emphasize these ranges differently from the CA3, and thus the weighting for these oscillations provide additional specificity for future work utilizing this tuning pipeline. All networks presented in this work achieved a score of 1.0 in all 3 frequency ranges indicating that this may assist in the speed to which each range is optimized, but little effect on the overall final tuning.

## Frequency Score

The third criterion, achieving biologically realistic mean firing rates, is quantified based on how similar the mean firing rate of each neuron type is relative to the target firing rate (Supplementary Fig. 3). A normalized probability density function centered on the full-scale firing rate thus gives a score from zero to one based on how close the firing rate of the tested network neuron is to the target frequency. Additionally, a standard deviation of the distribution of 1/10th the full-scale network firing rate was chosen after preliminary testing.

$$F(N) = \frac{1}{\#\, Groups} \sum_{g=0}^{\#\, Groups} PDF_{norm}(f_{Full,g}, 0.1 f_{Full,g}, f_{N,g}) \quad (4)$$

Where the $PDF_{norm}$ is the normalized probability density function with mean $f_{Full,g}$, the firing rate of group $g$ in the full-scale network, with standard deviation $0.1 f_{Full,g}$. This is used to calculate a value from zero to one of $f_{N,g}$, the firing rate of group $g$ in network $N$. This is then averaged across all groups by dividing by the number of groups.
Initial testing indicated that networks could tune several but not all neuron types to a high precision of firing rate. The simplicity of these scoring criteria allows for flexibility towards different potential use cases such as adjustments to the distribution standard deviation to allow for a larger window of acceptable values.

### Change in Connection Probability Score

The change in connection probability score is quantified similarly to the firing rate score. Optuna will control connection probabilities, given a range based on the biological values expected for a network of a given scale. Using a normalized PDF, with the target connection probability and standard deviation of 1/10$^{th}$ that value, a score for changes from the scaled connection probability is determined. Within a network, similar to the frequency score, each connection is weighed equally within the change in connection probability score. Optuna will try and maximize this score, which minimizes the changes (Supplementary Fig. 4). Because this change of scoring is based on the same principles as the firing rate scoring, the same considerations are available here as well.

## Downscaling Experimentation

Note that networks were tested in a range of configurations, notable 3 downscaling factors: 0.1, 0.01, 0.0001 (Fig. 2) and then downscaling the number of neuron types (Fig. 3). Each tuning included 2000 trials, representing 2000 different networks that would be generated and evaluated of a 5000 ms simulation duration. This automated process was conducted on a high-performance computing system to allow for a parallel trial process for different configuration parameters. After simulations were completed, the best trial by overall score was analyzed utilizing a variety of non-tuned metrics. Network synchronicity provides a numerical metric for the alignment of neurons within a population. Previous comparative work has indicated the Tiesinga-Sejnowski has the strongest correlation for determining synchronicity within single-scale spike trains[37,38].

$$S_{ts} = \frac{1}{\sqrt{N}} * \frac{\sqrt{\mu_{ISI^2} - {\mu_{ISI}}^2}}{\mu_{ISI}} \tag{6}$$

Where the Tiesinga-Sejnowski synchrony, $S_{ts}$, compares the squared mean of interspike intervals, ${\mu_{ISI}}^2$, with the mean of the squared interspike interval, $\mu_{ISI^2}$, and includes a term to account for neuron population size, $N$, to allow for comparison across subtypes and network sizes. Supplementary Fig. 8 provides example interspike interval distributions and their corresponding synchrony scores.

Beyond synchrony, the Gini coefficient provides us comparison to the literature and an analysis of per neuron activity equality across the network. Gini coefficients are calculated by summing activity on a per neuron basis across a given simulation. The total spikes per neuron defines the Lorenz curve of activity, and the Gini coefficient is the area between the line of equality and the Lorenz curve (Supplementary Fig. 9)[9].

Additional metrics on spike count and spiking transmission allow for analysis of potential energy consumption of these networks. A given spike propagation generates a communication event, that can be considered as a single spike or as a set of synaptic events from the one pre-synaptic neuron to one or more post-synaptic neurons. The total synaptic events, thus determines the total number of Accumulate Operations, ACC ops, the network is performing, and provides a basis for comparison to other algorithmic methods[43].

.

## Memristive Spiking Network Implementation

Translation to energy-efficient memristive hardware used the Daffodil prototyping platform to interface with a 20,000 device array[27]. This platform provides a python package, which we have expanded, to define spiking neural networks. This addition provides a similar interface to CARLsim6[33], dividing neuron populations into groups, and then creating connections between those groups. Using the same configuration file defined by Optuna during the tuning process, a network can be loaded up within the Daffodil platform[34]. Neurons were simulated within python utilizing the NumPy package for efficiency to run on the FPGA processor[44]. The 9-parameter Izhikevich neurons utilized the forward-euler method for simulation with the following equations[6]:

$$C_{mem}\frac{dv}{dt} = k(v - v_{rest})(v - v_{thres}) - u + I \tag{10}$$

$$\frac{du}{dt} = a\big(b(v - v_{rest})\big) - u \tag{11}$$

$$v > v_{peak} \rightarrow v = v_{reset} \tag{12}$$

$$v > v_{peak} \rightarrow u = u + d \tag{13}$$

Connections are processed through the memristive devices. This provides a precursor for networks utilizing the intrinsic properties of these devices for extremely efficient STDP[7,8]. For a given time step, each neuron is either spiking or silent. This corresponds to a read voltage, or a lower reference voltage set on the column lines of the crossbar array. Each of these read voltages generates a current through the devices corresponding to the connected postsynaptic current. This configuration allows for sending multiple spikes, and has the currents summed instantaneously across the row lines, allowing for extreme efficiency in synaptic calculations. However, a substantial portion of devices are in a stuck device state, meaning the device conductance is stuck, typically at an extremely high value, and unable to change. Because these devices, when read voltages are applied, generate currents beyond the expected range, a set of current readings can be done during network initialization to determine where stuck devices are. These currents can be stored, and while noisy, can be subtracted from subsequent current reads that involve that device. While this is a problem that needs to be addressed at the device fabrication level, this can be utilized here to create inherent sparsity in the network (since the currents generated are removed).

To mirror the biological synaptic calculations within CARLsim6, a conductance-based model was used[33]. The conductance follows an exponential decay function that receives an impulse based on the current measured from the row, indicating an incoming action potential. Each step was processed using the following:

$$I_{read} = (S * V_{read}) * G_{device} \quad (14)$$
$$I_{cleaned} = I_{read} - D_{stuck} * I_{base} \quad (15)$$
$$g_{impulse} = I_{cleaned} * \alpha \quad (16)$$
$$g_{syn} = g_{syn} * decay + g_{impulse} \quad (17)$$
$$excitatory: I_{syn} = g_{syn} * \left(0 - V_{post}\right) \quad (18)$$
$$inhibitory: I_{syn} = g_{syn} * \left(70 + V_{post}\right) \quad (19)$$

Where $I_{read}$ is the current read from each row, using read voltages, $V_{read}$, on every column, pre-synaptic neuron S, is firing. Then any devices that are in a stuck state removed to create $I_{cleaned}$. After a decay factor is multiplied to the synaptic conductance, the cleaned current is multiplied by a scaling factor and increments the synaptic conductance. This conductance is finally used with the corresponding reversal potential of excitatory and inhibitory neurons, 0mV and -70mV respectively.

An example 6 neuron network shows the transmission of activity from one group to another. Additionally, this example shows the ability for varying connection strengths to drive varying synaptic currents received by post-synaptic neurons (Fig 4a). All neuron equations are calculated and simulated within the processor on the FPGA board (Fig. 4b). These simulated neurons generate spikes which translate to hardware read voltages sent to the mix-signal daughterboard and through DACs and ADCs to communicate with the memristive devices. The memristive devices are organized into the 25x25 device kernels with DACs and ADCs (Fig. 4c). Device measurements across a range of target conductance levels and repeated individual device read operation. Normal distribution fits for each, read and write, indicate the near 0 mean error, with write errors tending toward a larger standard deviation than read errors (Fig. 4d). An example 5x5 subarray shows how a range of devices including devices with stuck current measurements since all read voltages are 0.3V through the DACs and current measurements from the ADCs (Fig. 4e-f). For a larger 12 neuron example comparison between CARLsim6, simulated, and physical hardware implementation see Supplementary Fig. 11.

## Mapping connection matrices

To efficiently simulate spiking neural networks on the mixed-signal platform, the synaptic connection matrix must be translated to sub-blocks of hardware devices. A one-to-one translation of synapses to device would require reading each device individually for each synaptic event. The hardware platform allows for up to 20,000 synapses given the 20,000 memristive devices across the entire chip. However, these are organized into 25x25 device crossbar limiting the number of devices we can concurrently read. This hardware architecture constrains how we can map the connection matrix of a given network. First, the networks with fewer 20,000 connections cannot be directly mapped with every possible presynaptic and postsynaptic combination per device and then adjust device conductance if there exists a connection in the network at that intersection. Only connections that exist between two neurons can be mapped to hardware given limited memristive devices. Second, to optimize the hardware efficiency during runtime, blocks of connections should be mapped together where one or more pre-synaptic neurons connect all-to-all with one or more post-synaptic neurons. This creates a single 'block' of connections that can be processed in a single read and is limited by kernel size of being up to 25 by 25 neurons in size. The efficiency of a device crossbar array comes from the matrix multiplication step that occurs through applying voltages across multiple columns, which are then multiplied by the conductivity of the devices and summed across the row of the array. Thus, devices need to be mapped to a corresponding set of synapses utilizing the same pre-synaptic neurons (sending spikes across the columns of the array) and the same post-synaptic neurons.

To do this, we have developed a greedy algorithm parameterized to take a set of neuron population sizes and connection probabilities and generate an efficient mapping for hardware implementation (Supplementary Fig. 12). Each group-to-group connection is processed separately, allowing for block parameter variation between different connection types. Given group sizes and a connection probability, random selections of N pre-synaptic and M post-synaptic neurons are generated. Each selection is compared to previous selections, if the number of connections within the randomly selected mapping exceeds a given threshold allowed, then the block is discarded and a new random selection is generated. Progressively the ability to randomly choose non-overlapping blocks decreases. If too many blocks are discarded in succession, we reduce the block size to reduce probability of overlapping, as well as reduce the threshold of non-overlapping connections (Fig. 4f-h and Supplementary Fig. 13). The final collection of NxM blocks are given locations within kernels. Figs. 4i-j show an example kernel on hardware and in simulation, while Supplementary Fig. 14 shows the full implementation.

Optimal NxM block sizing requires initial simulation testing to determine the tradeoff of computational efficiency of larger blocks with the reduced synaptic diversity that occurs with larger blocks. Initial trials maximizing block size create network behavior that lacks periodic behavior. Combinations of blocks sizes of varying shapes indicated that continuous periodic behavior is observed best when utilizing an Nx1 matrix, where multiple neurons are mapped to single post-synaptic neuron. Mapping to multiple post-synaptic neurons in a single block causes network activity to lose periodicity, which we hypothesize is due to the inherent over representation of a wider disparity in the number of outgoing connections a neuron could have and thus fewer neurons have disproportionately higher synaptic transmission. Additional preliminary testing indicated that a 12x1 matrix in combination with a 5x1 matrix for outgoing excitatory connection blocks provides a balance of efficient mapping and maintains periodic behavior.

# Data Availability

Data is available through Git Large File System (LFS) within the analysis repository and available at https://github.com/ADAM-Lab-GW/Downscaled-CA3-Tuning.

# Code Availability

Code is available with source control, analysis scripts, and documentation at https://github.com/ADAM-Lab-GW/Downscaled-CA3-Tuning.

# References

1. Klausberger, T. *et al.* Brain-state- and cell-type-specific firing of hippocampal interneurons in vivo. *Nature* **421**, 844–848 (2003).
2. Tremblay, R., Lee, S. & Rudy, B. GABAergic Interneurons in the Neocortex: From Cellular Properties to Circuits. *Neuron* **91**, 260–292 (2016).
3. Wheeler, D. W. *et al.* Hippocampome.org 2.0 is a knowledge base enabling data-driven spiking neural network simulations of rodent hippocampal circuits. *eLife* **12**, RP90597 (2024).
4. Kopsick, J. D., Kilgore, J. A., Adam, G. C. & Ascoli, G. A. Formation and retrieval of cell assemblies in a biologically realistic spiking neural network model of area CA3 in the mouse hippocampus. *J Comput Neurosci* https://doi.org/10.1007/s10827-024-00881-3 (2024) doi:10.1007/s10827-024-00881-3.
5. Sanchez-Aguilera, A. *et al.* An update to Hippocampome.org by integrating single-cell phenotypes with circuit function in vivo. *PLoS Biology* **19**, 1–28 (2021).
6. Izhikevich, E. M. *Dynamical Systems in Neuroscience: The Geometry of Excitability and Bursting*. (MIT Press, Cambridge, Mass, 2007).
7. Tecuatl, C., Wheeler, D. W., Sutton, N. & Ascoli, G. A. Comprehensive Estimates of Potential Synaptic Connections in Local Circuits of the Rodent Hippocampal Formation by Axonal-Dendritic Overlap. *J. Neurosci.* **41**, 1665–1683 (2021).
8. Moradi, K., Aldarraji, Z., Luthra, M., Madison, G. P. & Ascoli, G. A. Normalized unitary synaptic signaling of the hippocampus and entorhinal cortex predicted by deep learning of experimental recordings. *Commun Biol* **5**, 1–19 (2022).
9. Kopsick, J. D. *et al.* Robust Resting-State Dynamics in a Large-Scale Spiking Neural Network Model of Area CA3 in the Mouse Hippocampus. *Cogn Comput* **15**, 1190–1210 (2023).
10. Balasubramanian, V. Brain power. *Proceedings of the National Academy of Sciences* **118**, 1–3 (2021).

11. Patterson, D. *et al.* Carbon Emissions and Large Neural Network Training. *arXiv.org* https://arxiv.org/abs/2104.10350v3 (2021).

12. The new NeuroAI. *Nat Mach Intell* **6**, 245–245 (2024).

13. Hasler, J. & Marr, H. B. Finding a roadmap to achieve large neuromorphic hardware systems. *Front. Neurosci.* **7**, (2013).

14. Pehle, C. *et al.* The BrainScaleS-2 accelerated neuromorphic system with hybrid plasticity. *Frontiers in Neuroscience* **Volume 16-2022**, (2022).

15. van Albada, S. J. *et al.* Performance Comparison of the Digital Neuromorphic Hardware SpiNNaker and the Neural Network Simulation Software NEST for a Full-Scale Cortical Microcircuit Model. *Front. Neurosci.* **12**, (2018).

16. Mayr, C., Hoeppner, S. & Furber, S. SpiNNaker 2: A 10 Million Core Processor System for Brain Simulation and Machine Learning. Preprint at http://arxiv.org/abs/1911.02385 (2019).

17. Wang, F. *et al.* Neuromorphic Simulation of Drosophila Melanogaster Brain Connectome on Loihi 2. *arXiv.org* https://arxiv.org/abs/2508.16792v1 (2025).

18. Young, A. R., Dean, M. E., Plank, J. S. & S. Rose, G. A Review of Spiking Neuromorphic Hardware Communication Systems. *IEEE Access* **7**, 135606–135620 (2019).

19. Neftci, E. O. & Averbeck, B. B. Reinforcement learning in artificial and biological systems. *Nat Mach Intell* **1**, 133–143 (2019).

20. Basu, A., Frenkel, C., Deng, L. & Zhang, X. Spiking Neural Network Integrated Circuits: A Review of Trends and Future Directions. Preprint at https://doi.org/10.48550/arXiv.2203.07006 (2022).

21. Sharp, T., Galluppi, F., Rast, A. & Furber, S. Power-efficient simulation of detailed cortical microcircuits on SpiNNaker. *Journal of Neuroscience Methods* **210**, 110–118 (2012).

22. Lee, Y., Park, H.-L., Kim, Y. & Lee, T.-W. Organic electronic synapses with low energy consumption. *Joule* **5**, 794–810 (2021).

23. Uludağ, R. B., Çağdaş, S., İşler, Y. S., Şengör, N. S. & Aktürk, İ. Bio-realistic neural network implementation on Loihi 2 with Izhikevich neurons. *Neuromorph. Comput. Eng.* **4**, 024013 (2024).

24. Prezioso, M. *et al.* Spiking neuromorphic networks with metal-oxide memristors. in *2016 IEEE International Symposium on Circuits and Systems (ISCAS)* 177–180 (IEEE Press, Montréal, QC, Canada, 2016). doi:10.1109/ISCAS.2016.7527199.

25. Xia, Z. *et al.* Low-Power Memristor for Neuromorphic Computing: From Materials to Applications. *Nanomicro Lett* **17**, 217 (2025).

26. Yousuf, O. *et al.* Layer ensemble averaging for fault tolerance in memristive neural networks. *Nat Commun* **16**, 1250 (2025).

27. Hoskins, B. *et al.* A System for Validating Resistive Neural Network Prototypes. in *International Conference on Neuromorphic Systems 2021* 1–5 (Association for Computing Machinery, New York, NY, USA, 2021). doi:10.1145/3477145.3477260.

28. Ramirez-Morales, R. R. *et al.* Analog Implementation of a Spiking Neuron with Memristive Synapses for Deep Learning Processing. *Mathematics* **12**, 2025 (2024).

29. Romaro, C., Najman, F. A., Lytton, W. W., Roque, A. C. & Dura-Bernal, S. NetPyNE implementation and scaling of the Potjans-Diesmann cortical microcircuit model. *Neural Comput* **33**, 1993–2032 (2021).

30. Arroyo, J. I., Savage, V., Desai-Chowdhry, P., Kempes, C. & West, G. Scaling in Nervous Systems. https://ecoevorxiv.org/repository/view/7511/ (2024).

31. Cain, N., Iyer, R., Koch, C. & Mihalas, S. The Computational Properties of a Simplified Cortical Column Model. *PLOS Computational Biology* **12**, e1005045 (2016).

32. Morrison, A., Diesmann, M. & Gerstner, W. Phenomenological models of synaptic plasticity based on spike timing. *Biol Cybern* **98**, 459–478 (2008).

33. Niedermeier, L. *et al.* CARLsim 6: An Open Source Library for Large-Scale, Biologically Detailed Spiking Neural Network Simulation. in *2022 International Joint Conference on Neural Networks (IJCNN)* 1–10 (2022). doi:10.1109/IJCNN55064.2022.9892644.

34. Akiba, T., Sano, S., Yanase, T., Ohta, T. & Koyama, M. Optuna: A Next-generation Hyperparameter Optimization Framework. in *Proceedings of the 25th ACM SIGKDD International Conference on Knowledge Discovery & Data Mining* 2623–2631 (Association for Computing Machinery, New York, NY, USA, 2019). doi:10.1145/3292500.3330701.

35. Oswald, V. *et al.* Spontaneous brain oscillations as neural fingerprints of working memory capacities: A resting-state MEG study. *Cortex* **97**, 109–124 (2017).

36. Milstein, A. D., Tran, S., Ng, G. & Soltesz, I. Offline memory replay in recurrent neuronal networks emerges from constraints on online dynamics. *The Journal of Physiology* **601**, 3241–3264 (2023).

37. Tiesinga, P. H. E. & Sejnowski, T. J. Rapid Temporal Modulation of Synchrony by Competition in Cortical Interneuron Networks. *Neural Comput* **16**, 251–275 (2004).

38. Baroni, F. & Fulcher, B. D. Synchrony, oscillations, and phase relationships in collective neuronal activity: A highly comparative overview of methods. *PLoS Comput Biol* **21**, e1013597 (2025).

39. Kilgore, J. A., Kopsick, J. D., Ascoli, G. A. & Adam, G. C. Biologically-informed excitatory and inhibitory ratio for robust spiking neural network training. *Sci Rep* **15**, 24798 (2025).

40. Venkadesh, S., Komendantov, A. O., Wheeler, D. W., Hamilton, D. J. & Ascoli, G. A. Simple models of quantitative firing phenotypes in hippocampal neurons: Comprehensive coverage of intrinsic diversity. *PLOS Computational Biology* **15**, e1007462 (2019).

41. Colgin, L. L. Rhythms of the hippocampal network. *Nat Rev Neurosci* **17**, 239–249 (2016).

42. Trimper, J. B., Galloway, C. R., Jones, A. C., Mandi, K. & Manns, J. R. Gamma Oscillations in Rat Hippocampal Subregions Dentate Gyrus, CA3, CA1, and Subiculum Underlie Associative Memory Encoding. *Cell Reports* **21**, 2419–2432 (2017).

43. Yik, J. *et al.* The neurobench framework for benchmarking neuromorphic computing algorithms and systems. *Nat Commun* **16**, 1545 (2025).

44. Harris, C. R. *et al.* Array programming with NumPy. *Nature* **585**, 357–362 (2020).

# Acknowledgements

The authors acknowledge Osama Yousef for his assistance in setting up and documenting the memristive prototyping platform, and Nate Sutton for his assistance with setting up CARLsim.

The authors acknowledge the use of high-performance computing clusters, advanced support from the research technology services, and IT support at George Washington University.

# Funding

This work was supported by the Department of Energy Office of Science ASCR via the ECRP grant DE-SC0025567 (GW) and CRCNS collaborative effort under grant numbers DE-SC00023000 (GWU) and DE-SC0022998 (GMU). Additional support was provided by the Air Force Office of Scientific Research under grant number FA9550-23-1-0173 (GW).

# Author Contributions

J.A.K. and G.C.A. conceived the scientific concept. J.A.K. developed the network downscaling algorithm, and G.C.A., J.D.K. and G.A.A. helped refine it. J.A.K. and Z.A implemented the downscaling algorithm. J.A.K. developed the greedy algorithm for hardware translation and expanded the prototyping platform to support the implementation of spiking neural networks. J.A.K. set up and ran all simulations, collected data, and generated analysis with feedback from all authors. J.A.K. wrote the initial manuscript and all authors participated in co-editing. G.C.A. and G.A.A. supervised the project.

# Supplementary

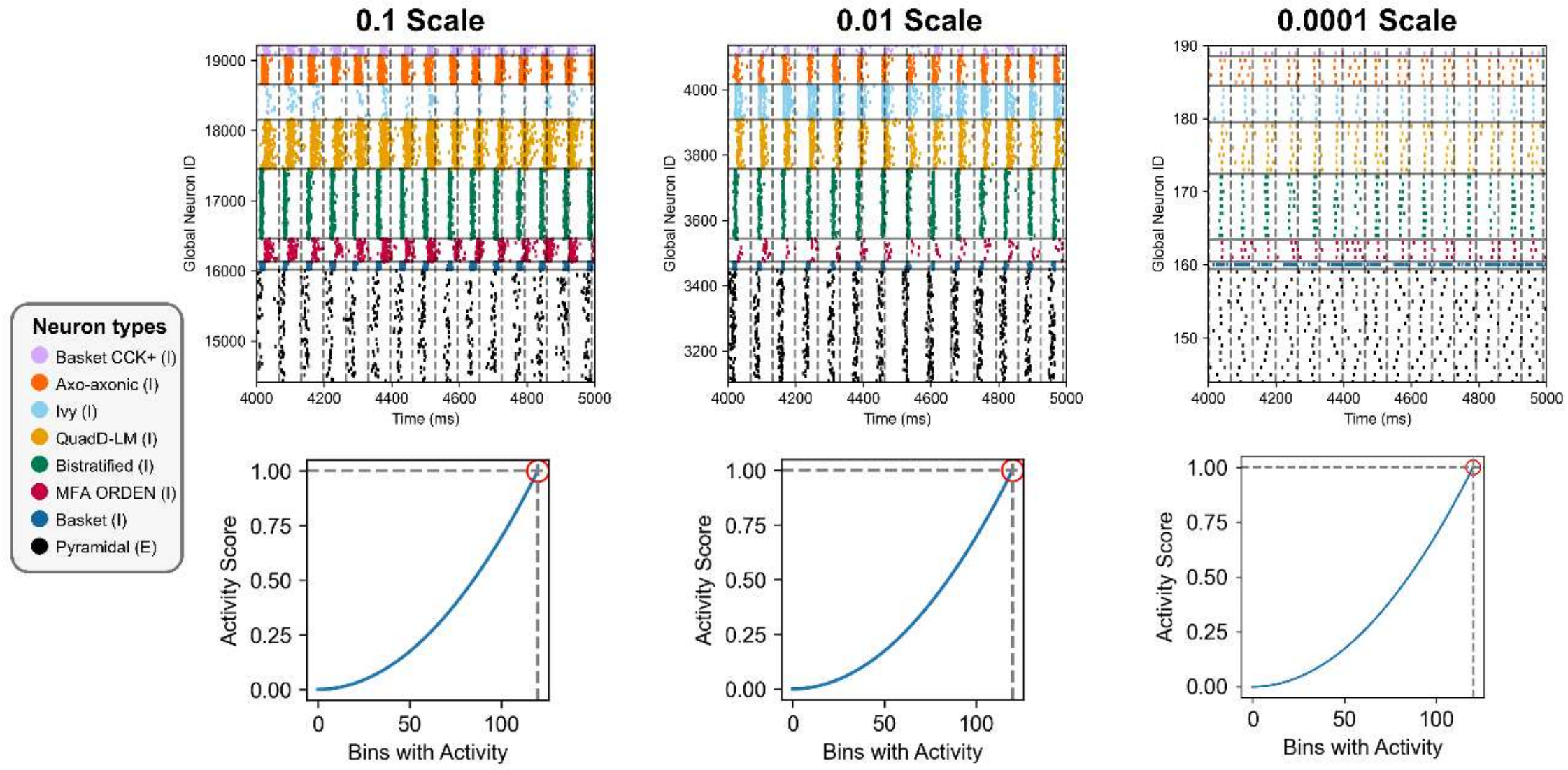

**Supplemental Figure 1.** Activity and related score for representative CA-inspired neural networks at 0.1, 0.01, and 0.0001 scales. Each network is divided into bins across time (indicated with vertical lines every 66ms) and across subtypes (indicated with horizontal lines between neuron types in the raster plots). The 66ms bins correspond to approximate overall frequency of 15Hz, noted in Kopsick et al. as the oscillation frequency of the full-scale network. The activity score follows a quadratic function, *A(N) Eq. (2)*, indicated with the blue line with no activity (0 bins with activity) scoring 0, and this example filling every bin with activity and having a perfect activity score of 1. The examples shown here all reach a perfect activity score since all neuron types have activity in all bins.

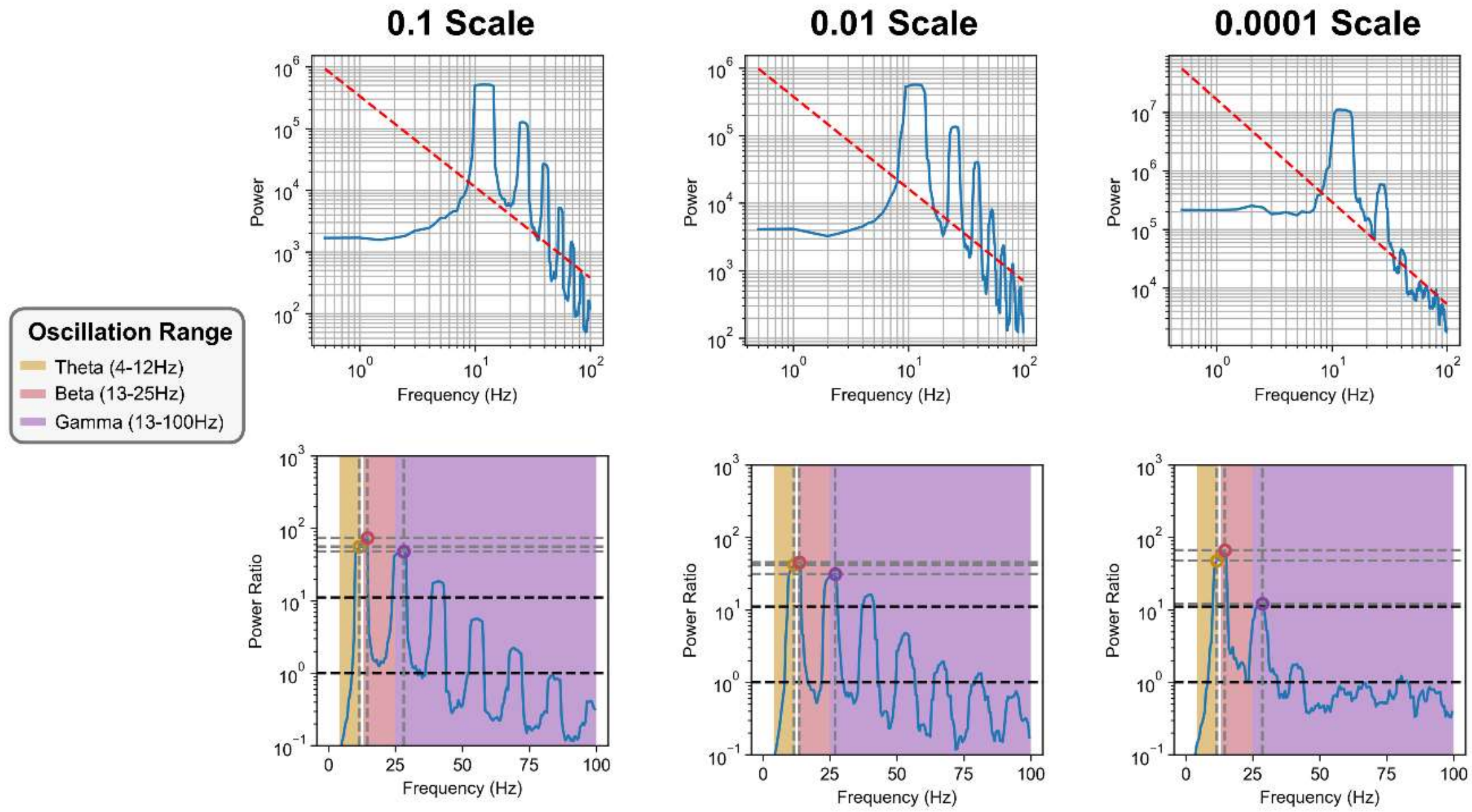

**Supplemental Figure 2.** Periodicity scores for the same representative neural networks at 0.1, 0.01, and 0.0001 scales as Supp. Fig. 2. Given the average voltage of the pyramidal cells, local field potential (LFP), a Fast Fourier Transform creates a power spectral density graph (top). Given this, a linear fit provides a baseline power level for a given frequency, from which peaks can quickly be determined. By dividing the LFP power spectral density by the line of best fit, the power ratio creates an immediate method for determining peak strength and comparing strengths of peaks across the frequency domain, with each peak indicated with a circle. Scores are then determined for each frequency range, theta (red), beta (green), and gamma (blue) where all scores above a power ratio of 11 (upper black horizontal line) score a full 1.0 for that frequency range periodicity score. Notably, all networks had a peak theta range power ratio at 11.5Hz, which is the upper end of the range. Conversely, networks tended toward a peak at the lower end of the beta oscillation range (e.g. 0.01 scale having a peak frequency of 13.5Hz).

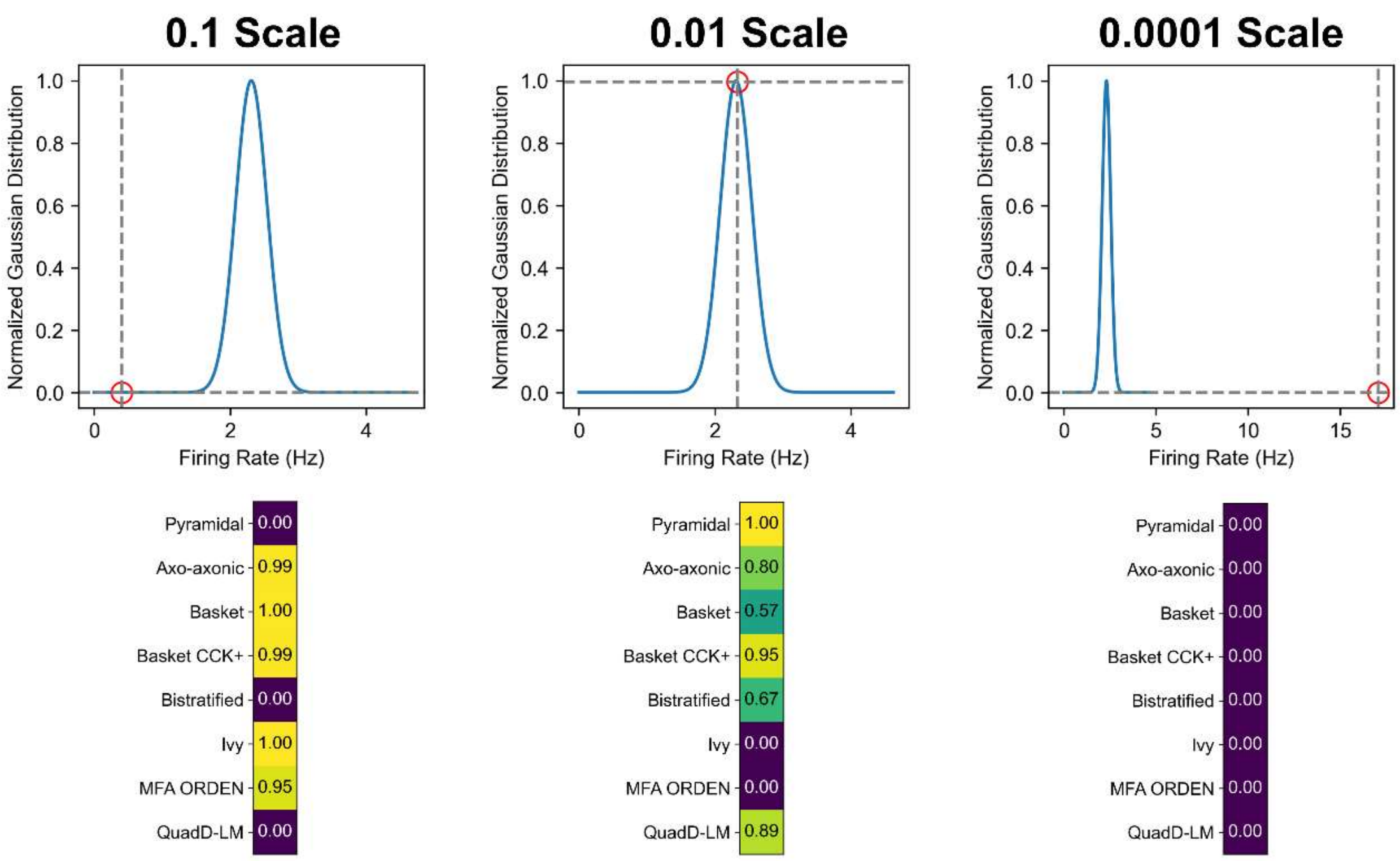

**Supplemental Figure 3.** Example scoring for the same representative neural networks at 0.1, 0.01, and 0.0001 scales as Supp. Figs. 1 & 2. The normal distribution shows the overall scoring curve for this neuron subtype, with each neuron having normal distribution with mean of the target firing rate and standard deviation of 1/10$^{th}$ of the mean. The 0.1 scale network produces 5 neuron types with scores ≥0.95. This reduces with scale, where the 0.01 scale network has only 2 neuron types ≥0.95, and the 0.0001 scale network having no neuron types near their target frequency.

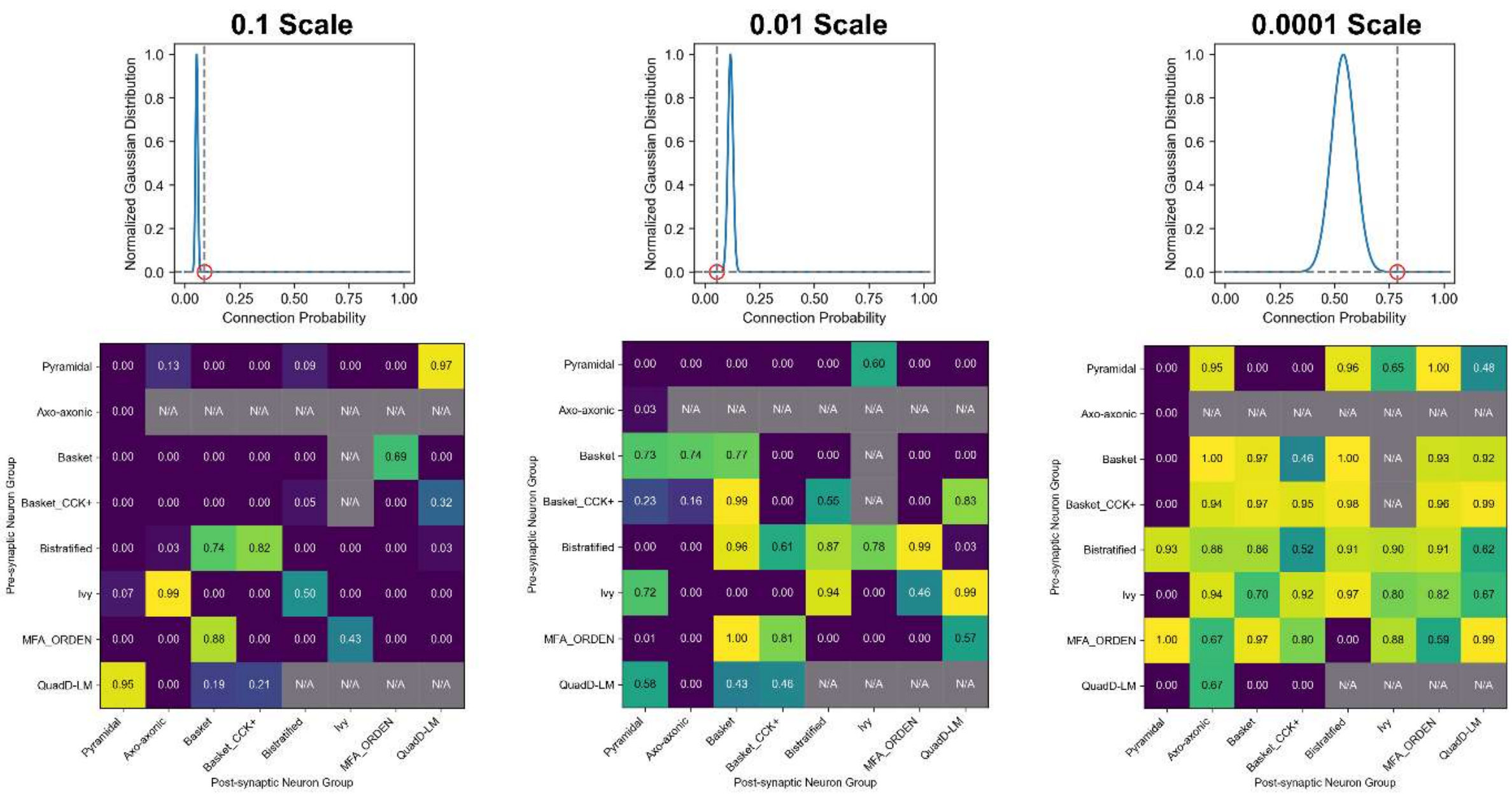

**Supplemental Figure 4.** The normalized Gaussian distribution defines the connection probability score for 0.1, 0.01, and 0.0001 scales. (top) Change score is based on the projected connection probability (the mean of the distribution) and $1/10^{th}$ the projected connection probability (the standard deviation of the distribution). Each connection generates a distinct curve and score, the example for pyramidal-to-pyramidal cells is shown. (bottom) Doing this across all connection types generates a heatmap for all of change scores that is averaged to produce the change score of the network. This 0.0001 scale example shows 15 connection probabilities at ≥0.95, substantially higher than the 3 connections in the 0.1 scale network.

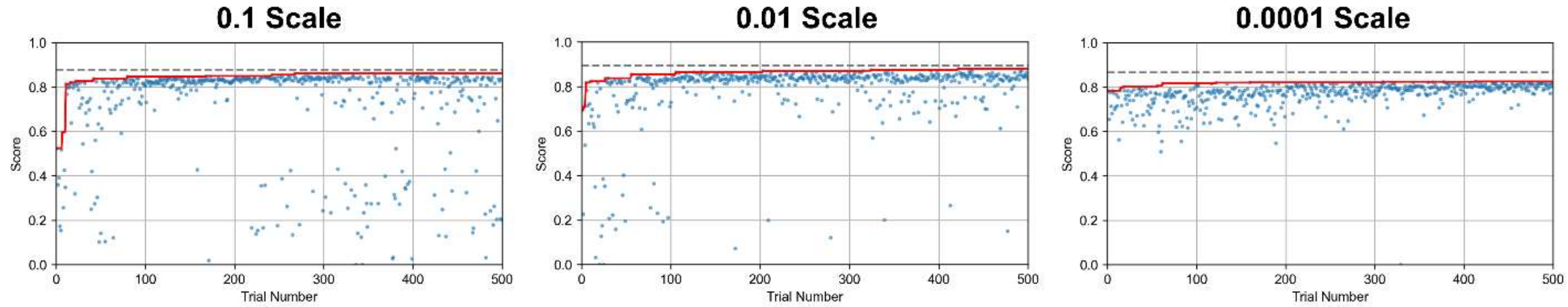

**Supplemental Figure 5.** First 500 trials of optimization convergence curves of 0.1, 0.01, and 0.0001 scale full diversity networks. Each trial represents an individual tested network configuration and the corresponding overall score that network generated. The best score achieved up to any given point is denoted with the red line. The dashed grey line indicates the highest score achieved across the 2000 trial tuning process with 0.1 scale achieving 0.9298, 0.01 scale achieving 0.8941, and 0.0001 scale achieving 0.8667.

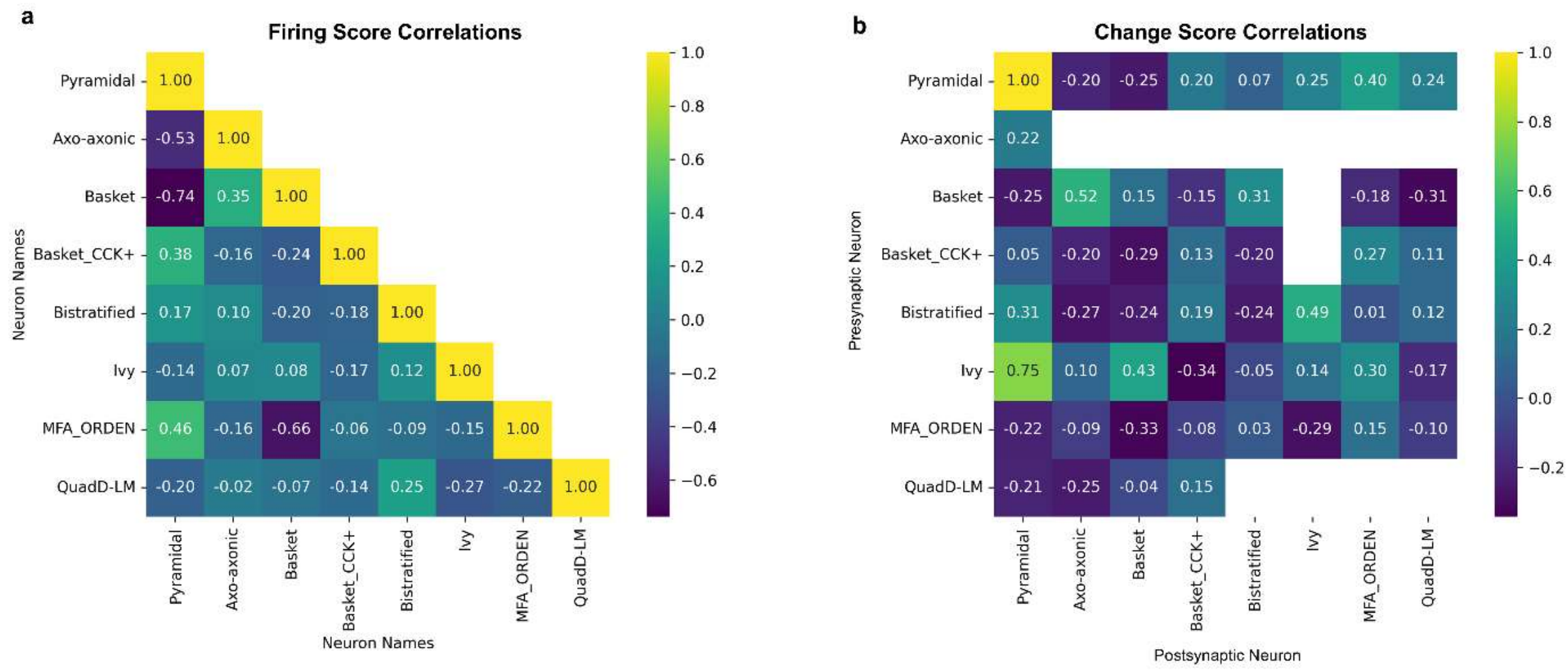

**Supplemental Figure 6.** Correlation matrix of firing rate scores and connection change scores across repeated trials at 0.001 scale. a) Correlation matrix of firing scores. Notably, the strongest correlation across neuron types was an inverse relationship between the pyramidal cell and the basket cell, with the basket and MFA ORDEN also having a high inverse correlation. Because the firing rate score purely indicates whether a neuron tuned to a biologically realistic firing rate, an inverse relationship indicates tradeoffs that occur when tuning. b) The correlation of the highest population connection, pyramidal to pyramidal, to all other connections. Here we again note a range of correlations, each with neuron types having positive and negative correlations with no distinct patterns. This may mean that the greater tradeoffs occurred with the firing rate scores, and that the connection probabilities, given the larger number, have greater flexibility with a highly diverse network.

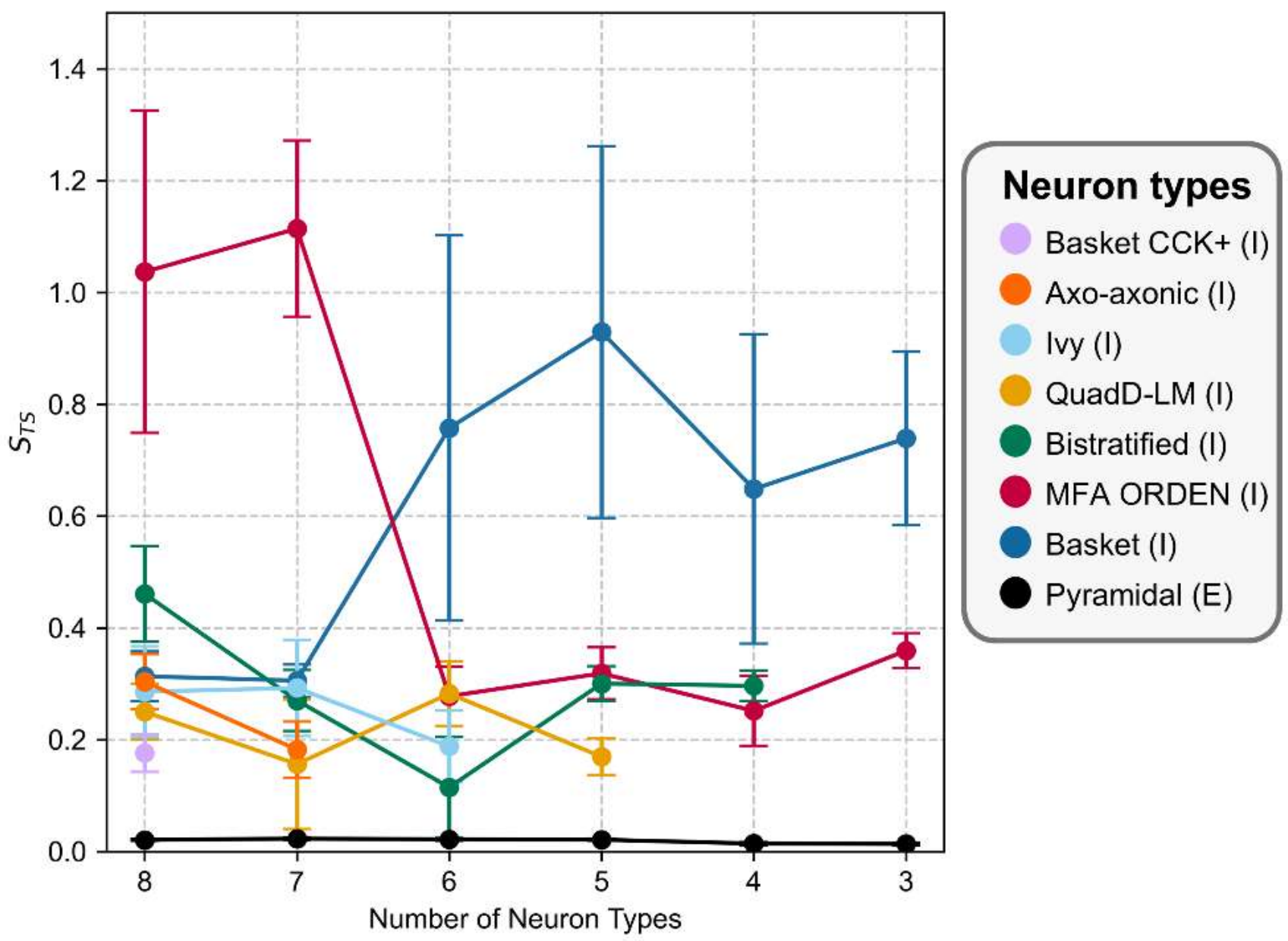

**Supplemental Figure 7.** Tiesinga-Sejnowski synchrony measures all neuron types at 0.0001 scale with reduced diversity. Across the diversity spectrum, one neuron type consistently had an average synchrony value across repeat trials above 0.6 while all other neuron types had an average synchrony below 0.5. However, this high synchrony neuron type switched from MFA ORDEN in 8 and 7 subtype networks to the basket cells in networks with 6 or fewer subtypes. The pyramidal cells are also consistently the lowest synchrony neuron type with the highest average $S_{TS}$ of 0.023 at 7 neuron types. The lowest average synchrony by any inhibitory neuron was 0.115 by the Basket CCK+ cell in 6 neuron type networks.

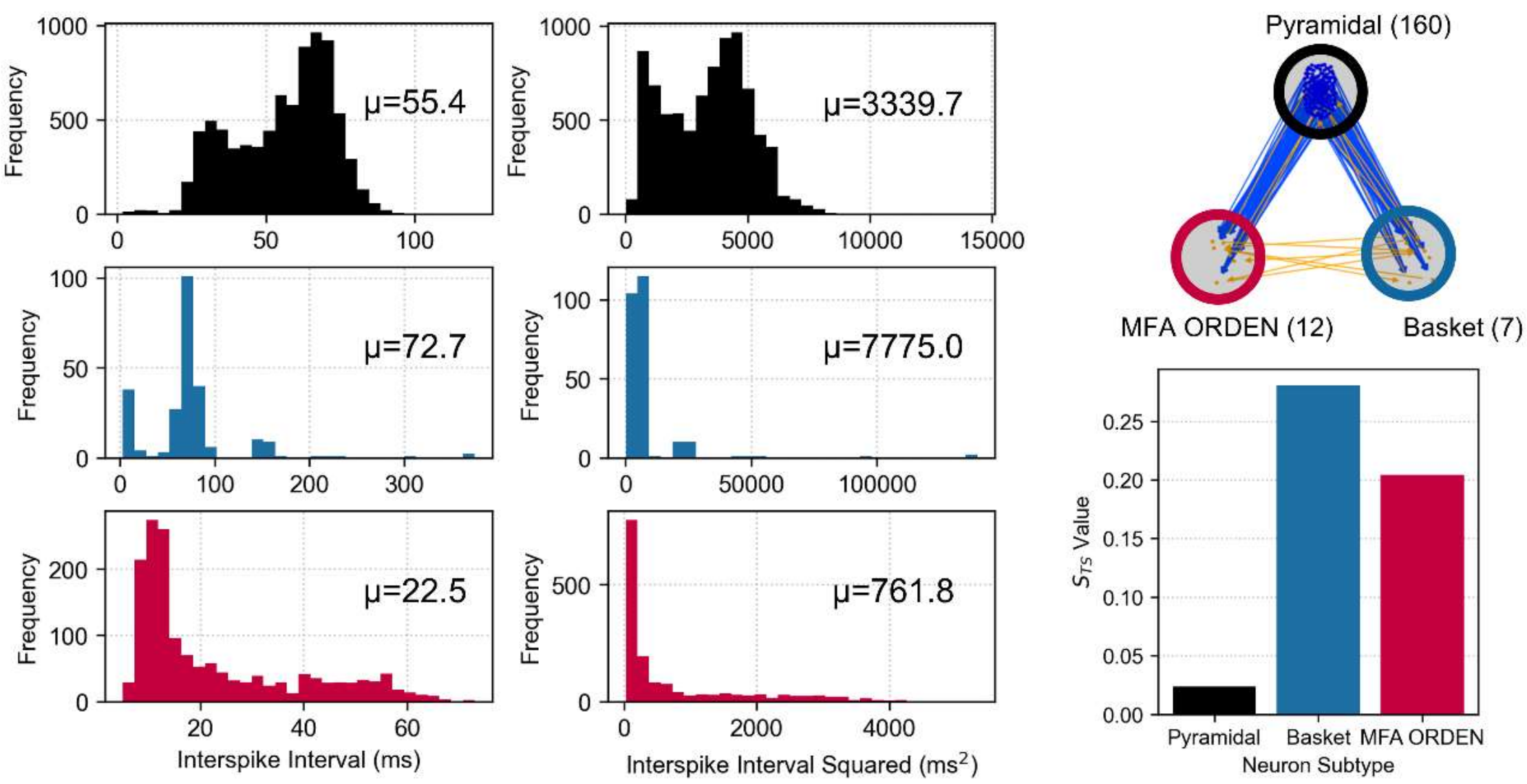

**Supplemental Figure 8.** The Tiesinga-Sejnowski synchrony measure is based on the interspike intervals and square of the interspike intervals for the physical hardware implementation. Thus, the two are shown, with the mean of each indicating the value used in the synchrony calculation. The resulting scores indicate that the basket cells, with a wide disparity of high and low interspike intervals, from 1 to >300ms, have the greatest synchrony at an $S_{TS}$ value of 0.28. Alternatively, the pyramidal cells, with a much larger frequency of interspike intervals occurring at the mean, 55.4ms, show a drastically lower synchrony score at 0.024.

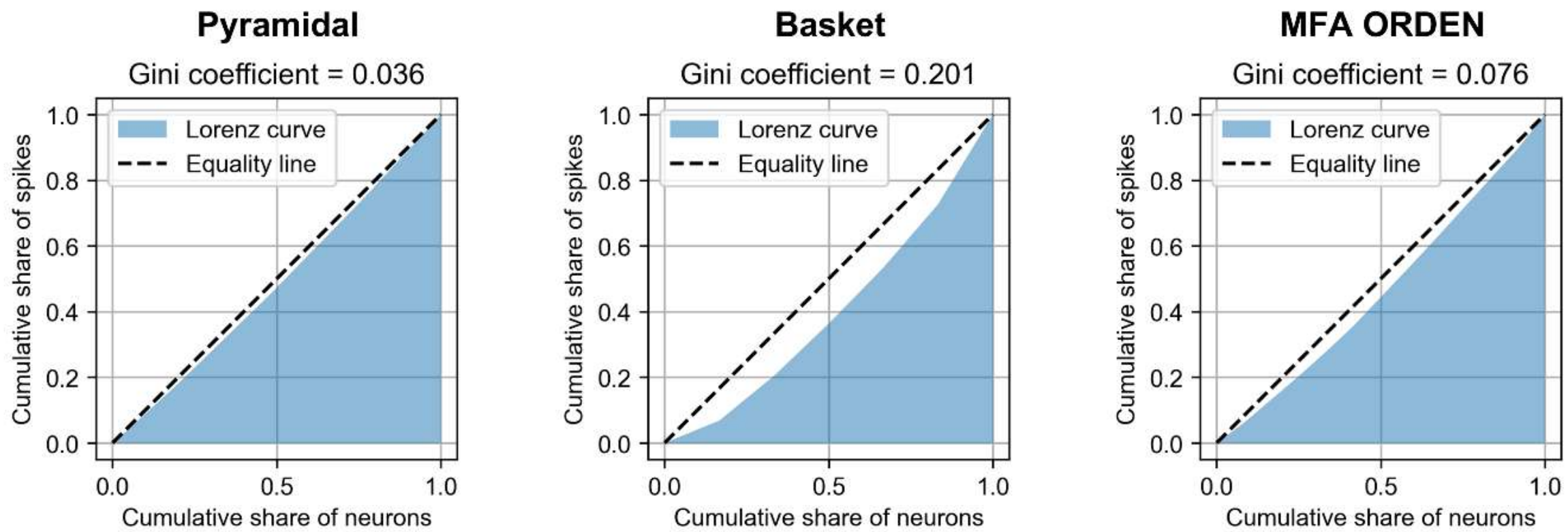

**Supplemental Figure 9.** Gini coefficient for each neuron subtype in 0.0001 scale reduced diversity network with physical memristive devices. Within each neuron type, the overall activity of each neuron can be accumulated across the population to create a unique Lorenz curve. The area between the equality line and the Lorenz curve defines the Gini coefficient. Here we note that all 3 neuron types show the pyramidal cells having high levels of equality, with MFA ORDEN cells the next more equally distributed. The basket cells are the most unequally distributed with a Gini coefficient of 0.201.

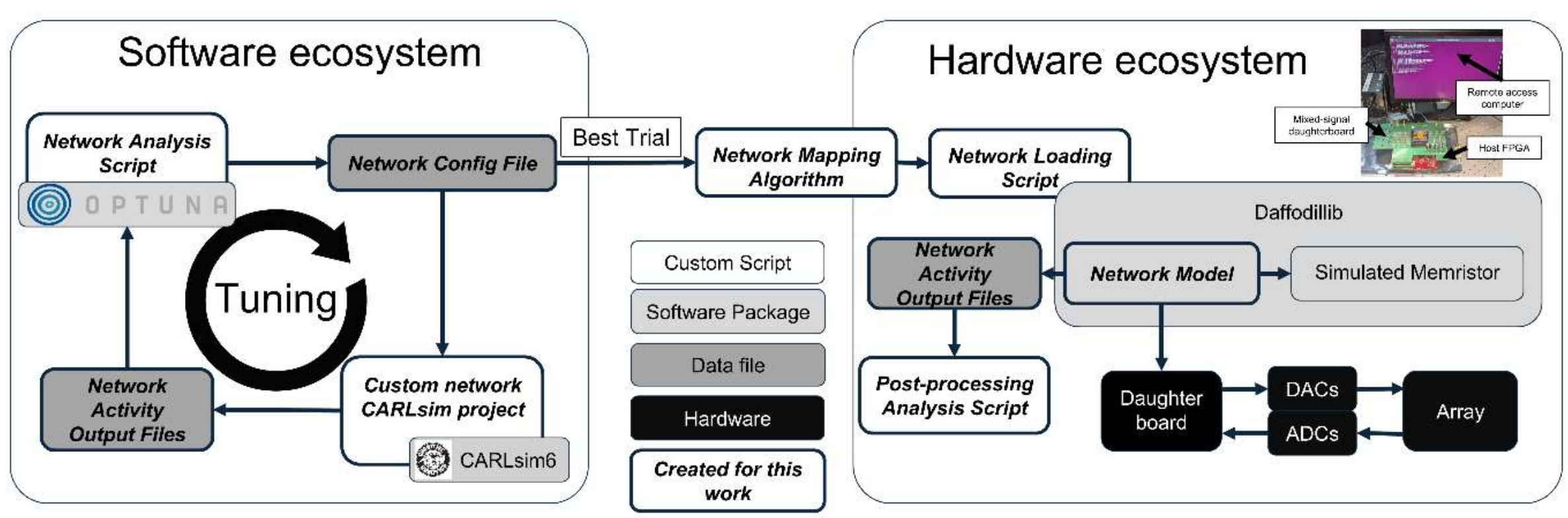

**Supplemental Figure 10.** The overall software-hardware prototyping ecosystem. The software ecosystem consists of 4 elements that feed into one another to support the tuning process. Initially the network analysis script, with the help of Optuna, generates a network configuration file. This configuration file is fed into a CARLsim6 executable and simulated with biologically realistic network dynamics, and the resulting activity is saved to an output file. Utilizing this output file the analysis script and a genetic algorithm, picks a new set of network parameters to run. This process continues as long as necessary, we typically used 5,000 trials per network variant. The resulting best trial configuration file is then processed through a separate mapping algorithm, transforming the data from sets of groups and connections that are needed to define a CARLsim simulation into a set of groups and connections that would map directly to hardware. This is then passed to a network loading script which supports the python wrapper for the hardware interface (called Daffodillib). This wrapper enables the network to run with either simulated or physical devices. The resulting activity is then extracted and passed to a separate post-processing analysis script that contains similar metrics to the network analysis script that started the tuning process.

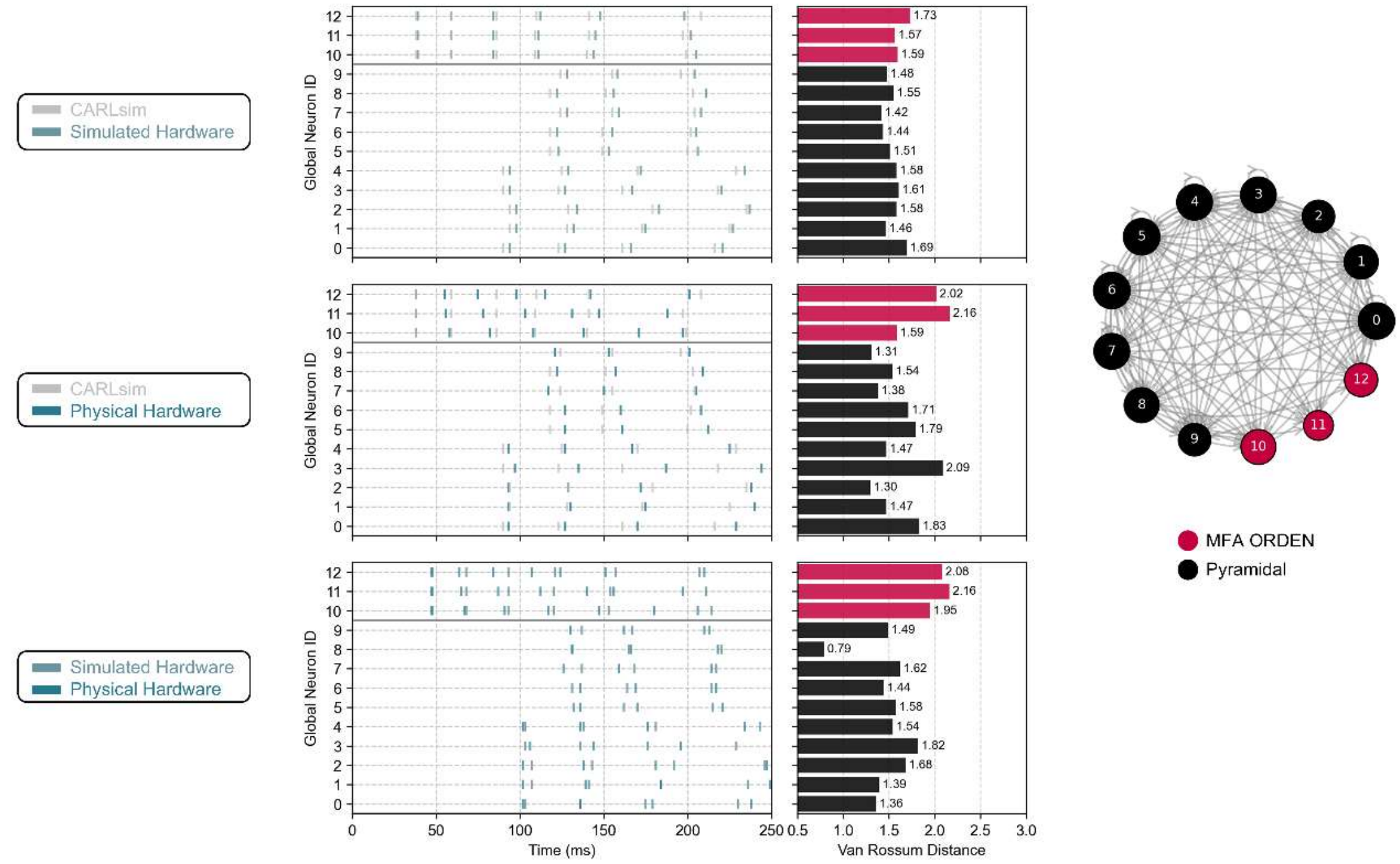

**Supplemental Figures 11.** Example toy network simulator comparison. This network contained 10 pyramidal cells and 3 MFA ORDEN cells. Each cell type was given a small input current, and the connections were mapped exactly for this toy network (containing ~100 connections). The Van Rossum distance provides a measure of distance for spike trains, comparing each neuron individually.

***Greedy Mapping:***

**Require**: Target connectivity $p$, between populations of size $N_{pre}$ and $N_{post}$. Initial 2-dimensional block size $k$ and a reduced block size $k_{fallback}$ with non-overlapping minimum of $o$ and overlapping relaxation factor of $o_{relax}$. A mapping matrix $M$ is initialized with size $N_{pre}$x$N_{post}$ and a list of mappings $S$ is initially empty. An attempt counter $a$ is initialized at 0, with a maximum attempts before condition relaxation of $a_{max}$.

**while** connectivity($M$) < $p$ **do**
    $n_{pre}$, $n_{post} \leftarrow$ set$_{random}$($N_{pre}$, $N_{post}$, $k$)
    $m \leftarrow$ ones($n_{pre}$, $n_{post}$)
    **if** 1-overlap($M$, $m$) < $o$ **then**
        $M \leftarrow M + m$
        $S \leftarrow S$ + ($n_{pre}$, $n_{post}$)
    **else**
        $a \leftarrow a + 1$
        **if** $a > a_{max}$ **then**
            $o \leftarrow o * o_{relax}$
            $k \leftarrow k_{fallback}$
            $a \leftarrow 0$
        **end if**
    **end if**
**end while**
**return** $S$

**Supplemental Figure 12.** Greedy mapping algorithm. The general process takes in a target connectivity, and generates blocks of connections, minimizing overlap, and repeating.

connectivity checks the connection probability overall of some matrix.

set$_{random}$($N_{pre}$, $N_{post}$, $k$) which selects $k$ values from $N_{pre}$, $N_{post}$, returning a matrix of pre- and post-synaptic neurons.

ones($n_{pre}$, $n_{post}$) generates a matrix of 1's for the entire shape of $n_{pre}$, $n_{post}$.

overlap($M$, $m$) checks the values of $m$ and calculates the percentage of $m$ that is in $M$ (i.e. there is a value of 1 in the same locations of $m$ and $M$).

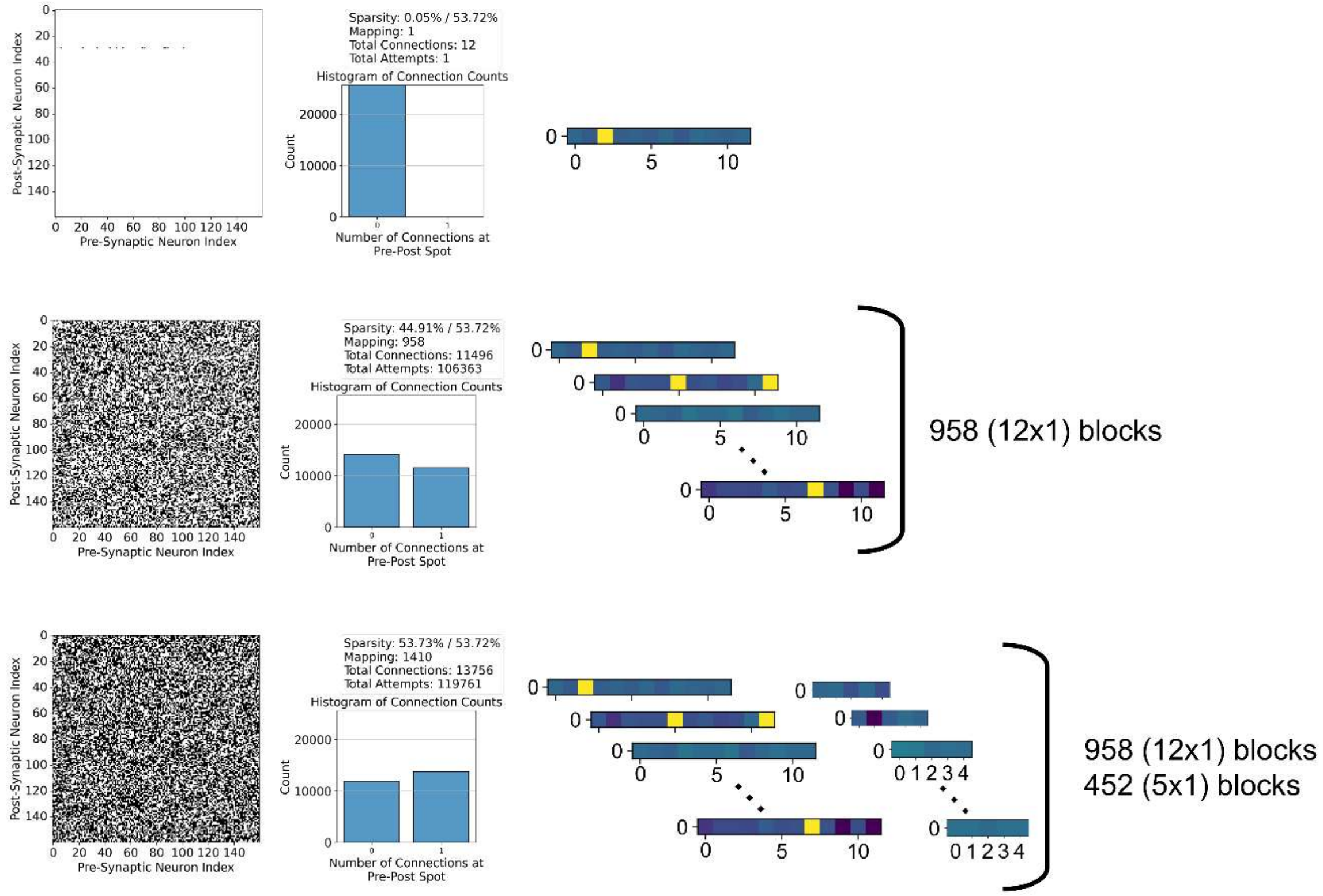

**Supplemental Figures 13.** Example blocking process for hardware mapping of the pyramidal-to-pyramidal connections. Here we start with a block size of 12x1, indicating 12 neurons connecting to a single neuron. This is repeated for the first 958 blocks, then changed to a 5x1 block size for the remainder of the mapping. In total, this approach generated 1,410 blocks for this connection type.

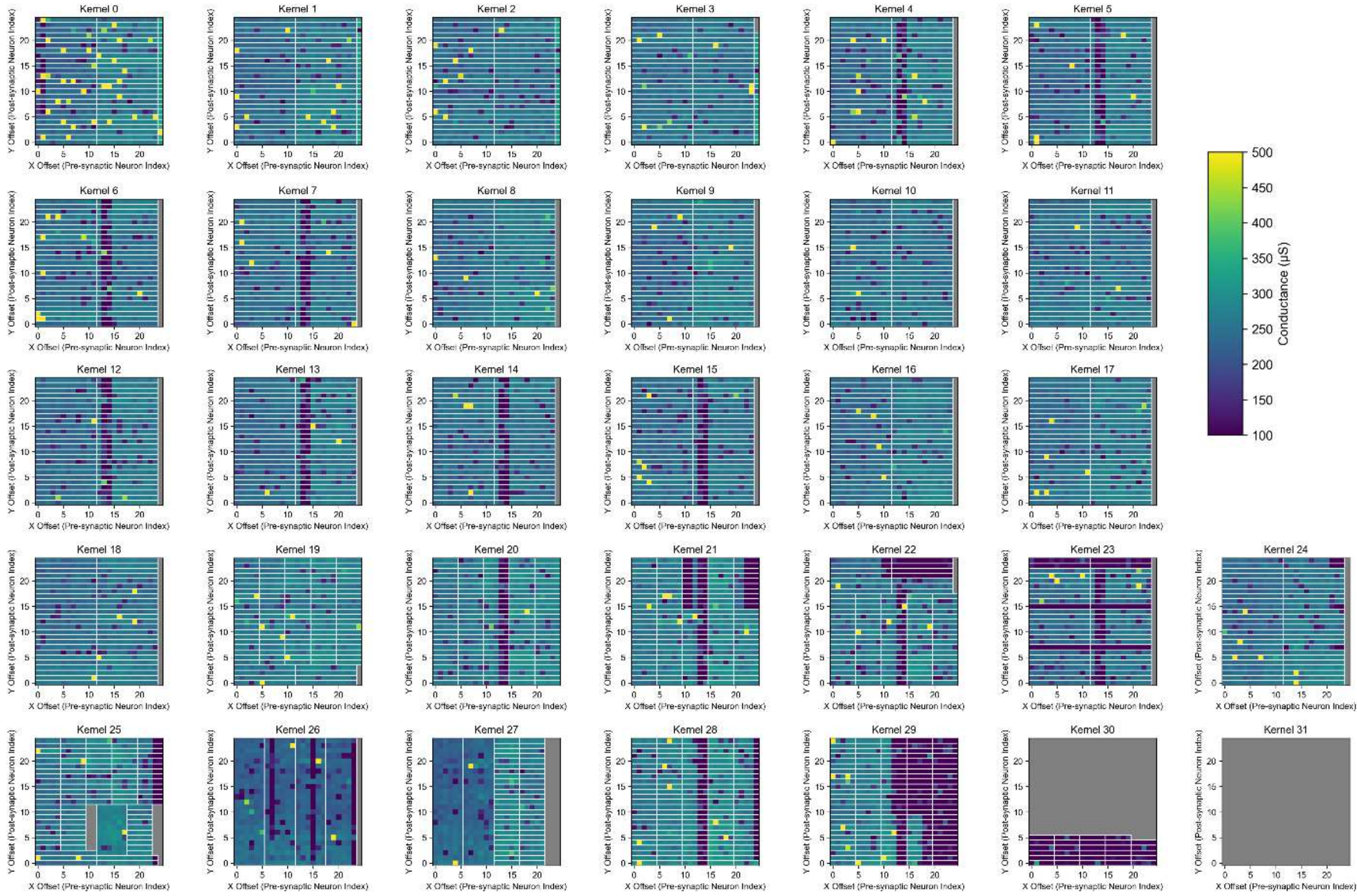

**Supplemental Figure 14.** The complete implementation of the 0.0001 CA3-inspired network on our prototyping platform with 20,000 memristive devices, organized in 32 kernels. During the initialization process of the network, and the allocation of physical devices to connection blocks, each device is measured individually to record an baseline conductance level. Each block is then stitched together to generate a baseline conductance heatmap for the entire network by individual kernel. Grey areas note devices that were not allocated during the mapping process, and white lines separate different connection blocks. Target conductance of 333μS was used, but not all devices reached this, and this initial measurement is used as part of the device current cleaning process in simulation during synaptic current calculation to be sent to the post-synaptic neuron.

**Supplemental Table 1** – Network Tuning

| Scale | Full-scale | 0.1 | 0.01 | 0.0001 | 0.0001** | 0.0001** | 0.0001 | 0.0001 | 0.0001 |
|---|---|---|---|---|---|---|---|---|---|
| Simulator | CARLsim with STDP | CARLsim | CARLsim | CARLsim | CARLsim | CARLsim | CARLsim | Simulated Hardware | Physical Hardware |
| Optuna Controlled Inhibitory Populations | N | N | N | N | N | N | Y | Y | Y |
| Neurons | 89,226 | 19,218 | 4,136 | 190 | 180 | 164 | 179 | 179 | 179 |
| Neuron Types | 8 | 8 | 8 | 8 | 5 | 3 | 3 | 3 | 3 |
| Connections | 250,078,223 | 32,275,304 | 1,652,658 | 26,597 | 21,581 | 11,474 | 17,996 | 18,316 | 18,316 |
| Connection Types | 51 | 51 | 51 | 51 | 22 | 9 | 9 | 9 | 9 |
| Percent Excitatory (Pyramdial) | 83.34% | 83.36% | 83.44% | 84.21% | 88.89% | 97.56% | 84.66% | 84.66% | 84.66% |
| Overall Score (1.0 max) | 1.0000 | 0.9298 | 0.8941 | 0.8667 | 0.8819 | 0.8726 | 0.9201 | 0.7807 | 0.8679 |
| Activity Score (0.4 max) | 0.4000 | 0.4000 | 0.4000 | 0.4000 | 0.4000 | 0.4000 | 0.4000 | 0.3516 | 0.4000 |
| Periodicity Score (0.4 max) | 0.4000 | 0.4000 | 0.4000 | 0.4000 | 0.3970 | 0.4000 | 0.4000 | 0.3462 | 0.3715 |
| Firing Rate Score (0.1 max) | 0.1000* | 0.0842 | 0.0611 | 0.0000 | 0.0188 | 0.0000 | 0.0333 | 0.0111 | 0.0230 |
| Change Score (0.1 max) | 0.1000 | 0.0456 | 0.0330 | 0.0667 | 0.0661 | 0.0751 | 0.0868 | 0.0718 | 0.0734 |
| Pyramidal $S_{TS}$ | 0.0051 | 0.0095 | 0.0080 | 0.0198 | 0.0135 | 0.0135 | 0.0208 | 0.0212 | 0.0235 |
| Basket $S_{TS}$ | 0.0373 | 0.0708 | 0.1364 | 1.0308 | 0.7593 | 0.7593 | 0.2496 | 0.3598 | 0.2806 |
| MFA ORDEN $S_{TS}$ | 0.0180 | 0.0399 | 0.0822 | 0.2632 | 0.3101 | 0.3101 | 0.2186 | 0.1842 | 0.2042 |
| Pyramidal Gini Coefficient | 0.8433 | 0.8619 | 0.1917 | 0.0235 | 0.0375 | 0.0158 | 0.0320 | 0.0368 | 0.0359 |
| Basket Gini Coefficient | 0.2261 | 0.1636 | 0.1335 | 0.0000 | 0.0000 | 0.0000 | 0.1784 | 0.3485 | 0.2008 |
| MFA ORDEN Gini Coefficient | 0.4048 | 0.2542 | 0.3830 | 0.0235 | 0.0561 | 0.0920 | 0.0344 | 0.0782 | 0.0765 |

*Firing rate score is based on CARLsim4 simulation firing rates, resulting in a perfect score, 0.1, for the previous CARLsim4 simulation available from Kopsick et al.
However utilizing CARLsim6 simulations, due to calculation updates, the resulting network has different firing rates and a correspoinding lower score, 0.0001.
**0.0001 scale with 5 or 3 neuron types and no optuna control for inhibitory neurons was run for 10 repeated trials.
The last trial is shown here as a choice representative of average behavior across trials.

**Supplemental Table 2** – Neuron Groups

CA3 Hippocampal Network — Neuron Group Parameters (Izhikevich)

| Neuron Type | E/I | Population Full-Scale | Target Firing Rate (Hz) | k | a | b | d | C (pF) | Vr (mV) | Vt (mV) | Vmin (mV) | Vpeak (mV) |
|---|---|---|---|---|---|---|---|---|---|---|---|---|
| Pyramidal | E | 74366 | 2.31 | 0.792 | 0.00838 | -42.552 | 588 | 366 | -63.204 | -33.604 | -38.868 | 35.861 |
| Axo-axonic | I | 1909 | 8.03 | 3.961 | 0.005 | 8.684 | 15 | 165 | -57.1 | -51.719 | -73.969 | 27.799 |
| Basket | I | 515 | 13.53 | 0.995 | 0.004 | 9.264 | -6 | 45 | -57.506 | -23.379 | -47.556 | 18.455 |
| Basket_CCK+ | I | 665 | 4.62 | 0.583 | 0.006 | -1.245 | 54 | 135 | -58.997 | -39.398 | -42.771 | 18.275 |
| Bistratified | I | 4631 | 6.9 | 3.935 | 0.002 | 16.58 | 19 | 107 | -64.673 | -58.744 | -59.703 | -9.929 |
| Ivy | I | 2334 | 2.36 | 1.916 | 0.008 | 12.933 | 45 | 364 | -70.435 | -40.859 | -53.4 | -6.92 |
| MFA_ORDEN | I | 1526 | 2.79 | 1.38 | 0.008 | 12.933 | 0 | 209 | -57.076 | -39.102 | -40.681 | 16.313 |
| QuadD-LM | I | 3280 | 7.24 | 1.776 | 0.006 | -3.449 | 52 | 186 | -73.482 | -54.937 | -64.404 | 7.066 |

**Supplemental Table 3** - Scaled Populations

| Subtype | Full-Scale | 0.1 | 0.01 | 0.0001 | 0.0001 (Hardware) |
|---|---|---|---|---|---|
| Pyramidal | 74366 | 16021 | 3451 | 160 | 160 |
| Axo-axonic | 1909 | 411 | 88 | 4 | |
| Basket | 515 | 110 | 23 | 1 | 7 |
| Basket_CCK+ | 665 | 143 | 30 | 1 | |
| Bistratified | 4631 | 997 | 214 | 9 | |
| Ivy | 2334 | 502 | 108 | 5 | |
| MFA_ORDEN | 1526 | 328 | 70 | 3 | 12 |
| QuadD-LM | 3280 | 706 | 152 | 7 | |
| **Total** | **89226** | **19218** | **4136** | **190** | **179** |

**Supplemental Table 4** – Connection Parameters

| Pre-Group Type | Post-Group Type | Prob. Conn. | Conductance | Delay (ms) | Range Low* | Range Med* | Range High* |
|---|---|---|---|---|---|---|---|
| Pyramidal | Pyramidal | 0.025 | 0.553 | 2 | 0 | 1.25 | 2.25 |
| Pyramidal | Axo-axonic | 0.015 | 0.874 | 2 | 0 | 0.7 | 1.7 |
| Pyramidal | Basket | 0.02 | 1.172 | 2 | 0 | 1.45 | 2.45 |
| Pyramidal | Basket_CCK+ | 0.017 | 0.848 | 2 | 0 | 1 | 2 |
| Pyramidal | Bistratified | 0.016 | 0.884 | 2 | 0 | 0.7 | 1.7 |
| Pyramidal | Ivy | 0.025 | 1.314 | 2 | 0 | 1.35 | 2.35 |
| Pyramidal | MFA_ORDEN | 0.021 | 0.88 | 2 | 0 | 1.25 | 2.25 |
| Pyramidal | QuadD-LM | 0.013 | 0.874 | 2 | 0 | 1.25 | 2.25 |
| Axo-axonic | Pyramidal | 0.15 | 1.87 | 1 | 0 | 1.45 | 2.45 |
| Basket | Pyramidal | 0.15 | 1.572 | 1 | 0 | 1.45 | 2.45 |
| Basket | Axo-axonic | 0.025 | 2.025 | 1 | 0 | 1.3 | 2.3 |
| Basket | Basket | 0.005 | 3.282 | 1 | 0 | 0.55 | 1.55 |
| Basket | Basket_CCK+ | 0.005 | 1.686 | 1 | 0 | 1 | 2 |
| Basket | Bistratified | 0.025 | 1.77 | 1 | 0 | 1.3 | 2.3 |
| Basket | MFA_ORDEN | 0.005 | 1.809 | 1 | 0 | 0.75 | 1.75 |
| Basket | QuadD-LM | 0.005 | 1.751 | 1 | 0 | 0.75 | 1.75 |
| Basket_CCK+ | Pyramidal | 0.15 | 1.306 | 1 | 0 | 1.45 | 2.45 |
| Basket_CCK+ | Axo-axonic | 0.025 | 1.494 | 1 | 0 | 1.3 | 2.3 |
| Basket_CCK+ | Basket | 0.005 | 1.745 | 1 | 0 | 0.55 | 1.55 |
| Basket_CCK+ | Basket_CCK+ | 0.005 | 0.966 | 1 | 0 | 1 | 2 |
| Basket_CCK+ | Bistratified | 0.025 | 1.371 | 1 | 0 | 1.3 | 2.3 |
| Basket_CCK+ | MFA_ORDEN | 0.005 | 1.355 | 1 | 0 | 0.75 | 1.75 |
| Basket_CCK+ | QuadD-LM | 0.025 | 1.335 | 1 | 0 | 0.75 | 1.75 |
| Bistratified | Pyramidal | 0.028 | 1.431 | 1 | 0 | 1.45 | 2.45 |
| Bistratified | Axo-axonic | 0.007 | 1.655 | 1 | 0 | 1.3 | 2.3 |
| Bistratified | Basket | 0.009 | 1.994 | 1 | 0 | 0.55 | 1.55 |
| Bistratified | Basket_CCK+ | 0.004 | 1.443 | 1 | 0 | 1 | 2 |

| Bistratified | Bistratified | 0.033 | 1.547 | 1 | 0 | 1.3 | 2.3 |
|---|---|---|---|---|---|---|---|
| Bistratified | Ivy | 0.004 | 2.061 | 1 | 0 | 0.65 | 1.65 |
| Bistratified | MFA_ORDEN | 0.009 | 1.568 | 1 | 0 | 1 | 2 |
| Bistratified | QuadD-LM | 0.008 | 1.49 | 1 | 0 | 0.75 | 1.75 |
| Ivy | Pyramidal | 0.072 | 1.541 | 1 | 0 | 1.45 | 2.45 |
| Ivy | Axo-axonic | 0.004 | 1.758 | 1 | 0 | 1.3 | 2.3 |
| Ivy | Basket | 0.016 | 2.111 | 1 | 0 | 0.55 | 1.55 |
| Ivy | Basket_CCK+ | 0.011 | 1.54 | 1 | 0 | 1 | 2 |
| Ivy | Bistratified | 0.017 | 1.66 | 1 | 0 | 1.3 | 2.3 |
| Ivy | Ivy | 0.004 | 2.143 | 1 | 0 | 0.65 | 1.65 |
| Ivy | MFA_ORDEN | 0.017 | 1.687 | 1 | 0 | 0.75 | 1.75 |
| Ivy | QuadD-LM | 0.002 | 1.567 | 1 | 0 | 0.75 | 1.75 |
| MFA_ORDEN | Pyramidal | 0.042 | 1.36 | 1 | 0 | 1.45 | 2.45 |
| MFA_ORDEN | Axo-axonic | 0.004 | 1.629 | 1 | 0 | 1.3 | 2.3 |
| MFA_ORDEN | Basket | 0.007 | 1.972 | 1 | 0 | 0.55 | 1.55 |
| MFA_ORDEN | Basket_CCK+ | 0.005 | 1.415 | 1 | 0 | 1 | 2 |
| MFA_ORDEN | Bistratified | 0.005 | 1.536 | 1 | 0 | 1.3 | 2.3 |
| MFA_ORDEN | Ivy | 0.003 | 2.082 | 1 | 0 | 0.65 | 1.65 |
| MFA_ORDEN | MFA_ORDEN | 0.002 | 1.553 | 1 | 0 | 0.75 | 1.75 |
| MFA_ORDEN | QuadD-LM | 0.004 | 1.472 | 1 | 0 | 0.75 | 1.75 |
| QuadD-LM | Pyramidal | 0.119 | 1.183 | 1 | 0 | 1.45 | 2.45 |
| QuadD-LM | Axo-axonic | 0.005 | 1.473 | 1 | 0 | 1.3 | 2.3 |
| QuadD-LM | Basket | 0.067 | 1.815 | 1 | 0 | 0.55 | 1.55 |
| QuadD-LM | Basket_CCK+ | 0.05 | 1.308 | 1 | 0 | 1 | 2 |

*Range Low, Med, and High are used to set an internal weight variable within CARLsim,
and are multiplied by the conductance during a synaptic event to calculate internal syanpstic conductance change